\documentclass{article}

\usepackage[final]{neurips_2022}
\usepackage[T1]{fontenc}
\usepackage[utf8]{inputenc}
\usepackage{microtype}
\usepackage{amsmath}
\usepackage{amssymb}
\usepackage{dsfont}
\usepackage{graphicx}
\usepackage{caption}
\usepackage{subcaption}
\usepackage{xcolor}
\usepackage[noend]{algpseudocode}
\usepackage{algorithm}
\usepackage{tabularx}
\usepackage{booktabs}
\usepackage{xcolor}
\usepackage{xcolor,mdframed}
\usepackage{hyperref}

\usepackage{color,soul}
\usepackage{tikz}
\usepackage{wrapfig}
\usepackage{microtype}
\usepackage{textcomp}
\usepackage{comment}
\usepackage[colorinlistoftodos,prependcaption,textsize=tiny]{todonotes}
\usepackage{xargs}
\usepackage{url}
\usepackage{enumitem}
\usepackage{makecell}
\usepackage{amsthm}

\newcommand{\E}{\mathbb{E}}

\newcommand{\KL}{D_{\mathrm{KL}}}
\newcommand{\KLhat}{\hat{D}_{\mathrm{KL}}}
\newcommand{\TVD}{\mathrm{TVD}}

\newcommand{\pit}{{\pi_\theta}}
\newcommand{\nablapitlog}{\nabla_{\theta} \log \pit(x)}

\newcommand{\Rtz}{\Rt} 
\newcommand\numberthis{\addtocounter{equation}{1}\tag{\theequation}}

\DeclareMathOperator*{\argmax}{arg\,max}
\DeclareMathOperator*{\argmin}{arg\,min}

\newcommand{\GDC}{GDC }

\newcommand{\GDCplus}{GDC\texttt{++} }
\newcommand{\DPG}{DPG }

\newcommand{\DPGoff}{DPG\textsuperscript{off} }

\newcommand{\Boff}{B^\text{off}(x)}

\newcommand{\grad}{G_\theta}
\newcommand{\gradest}{G(\theta)}
\newcommand{\vargrad}{\mathrm{Var}(\grad)}
\newcommand{\mgrad}{\mathbf{\mu}(\grad)}

\newcommand{\Adv}{\mathrm{A}}
\newcommand{\mAdv}{\mu^\Adv}
\newcommand{\mAbsAdv}{\mu^{|\Adv|}}
\newcommand{\varAdv}{\mathrm{Var}\left(\Adv\right)}

\newcommand{\nabt}{\nabla_\theta}
\newcommand{\Rt}{R_\theta}

\newcommand{\EX}[1]{\E_{#1}}

\def\Algref#1{Algorithm~\ref{#1}}

\newif\ifready

\readytrue

\ifready
\newcommand{\ftk}[1]{}
\newcommand{\fgk}[1]{}
\newcommand{\fmd}[1]{}
\newcommand{\ptmd}[1]{#1}
\newcommand{\ptgk}[1]{#1}
\newcommand{\fhe}[1]{}
\newcommand{\annot}[1]{}
\else
\newcommand{\ftk}[1]{{\color{red}\footnote{\textcolor{red}{TK: #1}}}}
\newcommand{\fgk}[1]{{\color{orange}\footnote{\textcolor{orange}{GK: #1}}}}
\newcommand{\fmd}[1]{{\color{blue}\footnote{\textcolor{blue}{MD: #1}}}}
\newcommand{\ptmd}[1]{\textcolor{blue}{#1}}
\newcommand{\ptmd}[1]{\textcolor{orange}{#1}}
\newcommand{\fhe}[1]{{\color{olive}\footnote{\textcolor{olive}{HE: #1}}}}
\newcommand{\annot}[1]{\textbf{\textcolor{purple}{#1~}}}
\fi

\newif\ifreadyII

\readyIItrue

\ifreadyII
\newcommand{\ftkII}[1]{}
\newcommand{\fgkII}[1]{}
\newcommand{\fmdII}[1]{}
\newcommand{\ptmdII}[1]{#1}

\newcommand{\fheII}[1]{}
\newcommand{\annotII}[1]{}
\else
\newcommand{\ftkII}[1]{{\color{red}\footnote{\textcolor{red}{TK: #1}}}}
\newcommand{\fgkII}[1]{{\color{orange}\footnote{\textcolor{orange}{GK: #1}}}}
\newcommand{\fmdII}[1]{{\color{blue}\footnote{\textcolor{blue}{MD: #1}}}}
\newcommand{\ptmdII}[1]{\textcolor{blue}{#1}}

\newcommand{\fheII}[1]{{\color{olive}\footnote{\textcolor{olive}{HE: #1}}}}
\newcommand{\annotII}[1]{\textbf{\textcolor{purple}{#1~}}}
\fi

\newif\ifreadyAR

\readyARtrue

\ifreadyAR
\newcommand{\ftkAR}[1]{}
\newcommand{\fgkAR}[1]{}
\newcommand{\fmdAR}[1]{}
\newcommand{\ptmdAR}[1]{#1}

\newcommand{\fheAR}[1]{}
\newcommand{\annotAR}[1]{}
\else
\newcommand{\ftkAR}[1]{{\color{red}\footnote{\textcolor{red}{TK: #1}}}}
\newcommand{\fgkAR}[1]{{\color{orange}\footnote{\textcolor{orange}{GK: #1}}}}
\newcommand{\fmdAR}[1]{{\color{blue}\footnote{\textcolor{blue}{MD: #1}}}}
\newcommand{\ptmdAR}[1]{\textcolor{blue}{#1}}

\newcommand{\fheAR}[1]{{\color{olive}\footnote{\textcolor{olive}{HE: #1}}}}
\newcommand{\annotAR}[1]{\textbf{\textcolor{purple}{#1~}}}
\fi

\newcommand{\Rtheta}{R_\theta}
\renewcommand{\Rtz}{R^z_\theta}
\newcommand{\Rpiz}{R^z_\pi}

\readyfalse

\newcommand{\smallparagraph}[1]{\textbf{#1}\hspace{1ex}}
\definecolor{g0}{HTML}{30b52d}
\definecolor{y1}{HSB}{38,10,255}
\definecolor{y2}{HSB}{38,20,255}
\definecolor{y3}{HSB}{38,25,255}
\definecolor{y4}{HSB}{38,50,255}
\definecolor{y5}{HSB}{38,80,255}
\definecolor{y6}{HSB}{38,110,255}
\definecolor{y7}{HSB}{38,150,255}
\definecolor{y8}{HSB}{38,200,255}
\definecolor{y9}{HSB}{38,250,255}
\definecolor{y10}{HSB}{38,255,255}
\definecolor{r0}{RGB}{255,181,195}
\definecolor{green_eq}{HTML}{006400}
\definecolor{orange_eq}{HTML}{CC6400}
\definecolor{gray_custom}{HTML}{404040}

\DeclareRobustCommand{\ye}[1]{{\sethlcolor{y5}\hl{#1}}}

\DeclareRobustCommand{\yg}[1]{{\sethlcolor{y7}\hl{#1}}}
\DeclareRobustCommand{\yh}[1]{{\sethlcolor{y8}\hl{#1}}}

\DeclareRobustCommand{\g}[1]{{\sethlcolor{g0}\hl{#1}}}
\DeclareRobustCommand{\r}[1]{{\sethlcolor{r0}\hl{#1}}}
\DeclareRobustCommand{\gc}[1]{{\color{gray_custom}{#1}}}

\newtheorem{theorem}{Theorem}
\newtheorem{fact}{Fact}

\title{On Reinforcement Learning and Distribution Matching for Fine-Tuning Language Models \\with no Catastrophic Forgetting}

\author{
Tomasz Korbak\thanks{Work partly done during an internship at Naver Labs Europe.} \\
University of Sussex \\
\texttt{tomasz.korbak@gmail.com}
\And
Hady Elsahar\\
Naver Labs Europe\\
\texttt{hady.elsahar@gmail.com}
\AND
Germán Kruszewski\\
Naver Labs Europe\\
\texttt{german.kruszewski@naverlabs.com}
\And
Marc Dymetman\thanks{Independent Researcher. Work done at Naver Labs Europe.}\\
\texttt{marc.dymetman@gmail.com}
}

\date{}

\begin{document}
\maketitle
\begin{abstract}
The availability of large pre-trained models is changing the landscape of Machine Learning research and practice, moving from a ``training from scratch'' to a ``fine-tuning'' paradigm. 
While in some applications the goal is to ``nudge'' the pre-trained distribution towards preferred outputs, in others it is to steer it towards a different distribution over the sample space.
Two main paradigms have emerged to tackle this challenge: Reward Maximization (RM) and, more recently, Distribution Matching (DM).
RM applies standard Reinforcement Learning (RL) techniques, such as Policy Gradients, to gradually increase the reward signal.
DM prescribes to first make explicit the target distribution that the model is fine-tuned to approximate.
Here we explore  
the theoretical
connections between the two paradigms, and show that methods such as KL-control developed 
for RM
can also be construed as belonging to DM.
We further observe that while DM differs from RM, it can suffer from similar training difficulties, such as high gradient variance.
We leverage connections between the two paradigms to import the concept of \textit{baseline} into DM methods.
We empirically validate the benefits of adding a baseline on an array of controllable language generation tasks such as constraining topic, sentiment, and gender distributions in texts sampled from a language model. We observe superior performance in terms of constraint satisfaction, stability and sample efficiency.

\end{abstract}

\section{Introduction}

Pre-trained language models \citep{devlin-etal-2019-bert,radford2019language} are changing the landscape of Machine Learning research and practice. 
Due to their strong generative capabilities many studies have found it sufficient to ``nudge'' these models to conform to global preferences defined over the generated sequences instead of training from scratch using annotated data. These preferences could include
topic and sentiment~\citep{plug_and_play_20}, valid musical notes and molecular structures~\citep{KL_Jaques17}, code compilability~\citep{korbak2021code}, reducing distributional biases~\citep{khalifa_2021, ethical_lm_deepmind}, evaluation metrics for Machine Translation and Summarization~\citep{seq_lvl_train_RanzatoCAZ15,bahdanau_actor-critic_2016}, or direct human feedback~\citep{Ziegler19,ziegler_summarize2021}.
This large body of studies is driven by two different paradigms: \emph{Reward Maximization} (RM) and \emph{Distribution Matching} (DM).

\begin{figure}
    \centering
    \includegraphics[width=0.9\linewidth]{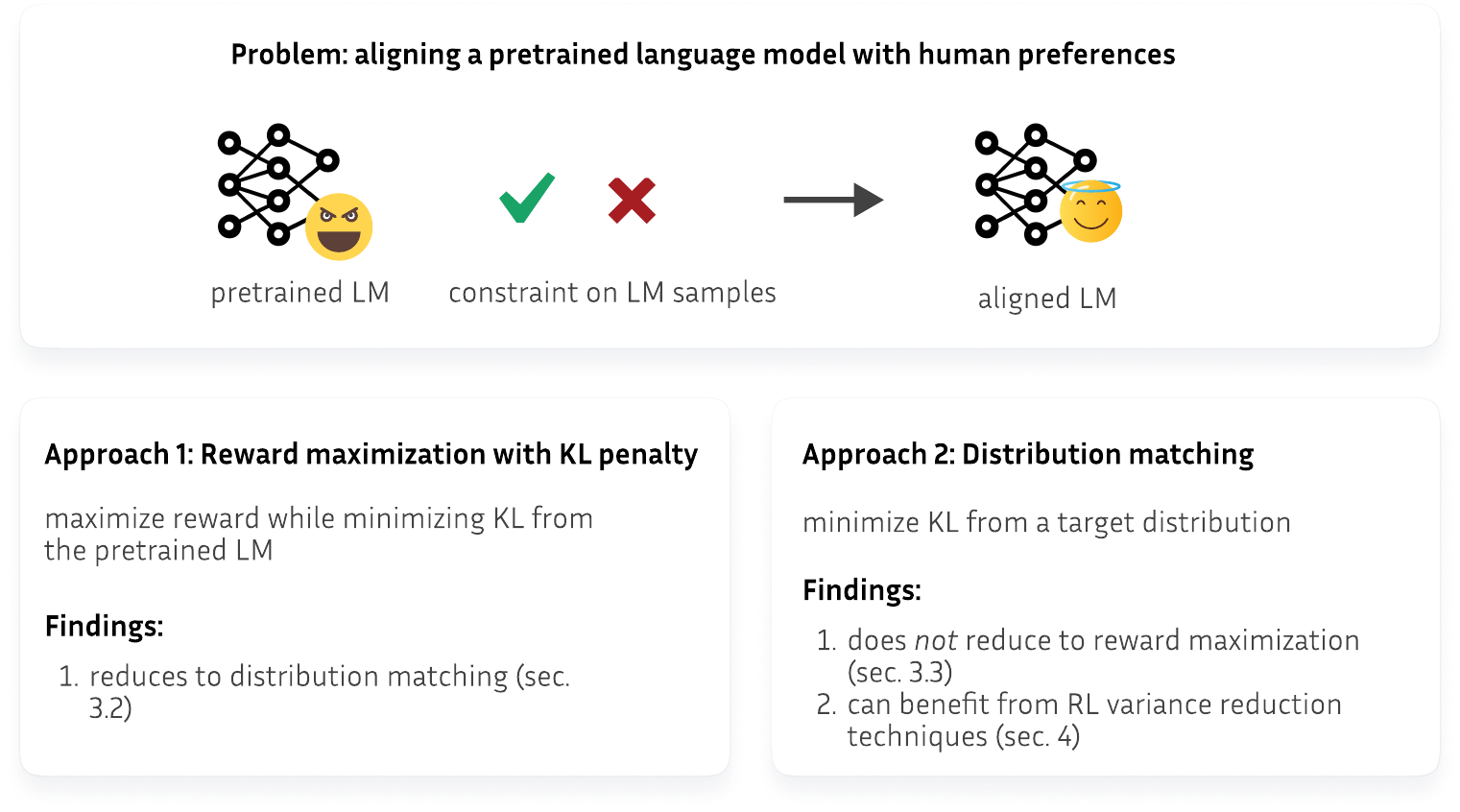} 
    \caption{\small{
    In this study we make a connection between two popular paradigms for aligning language models to human preferences: Reward maximization (RM) and Distribution matching (DM).
    }}
    \label{fig:intro}
\end{figure}

\paragraph{Reward Maximization}
RM intuitively nudge\ptmdII{s} pre-trained models towards \ptmdII{certain} preferences by providing global sequence-level rewards when the model generates outputs that satisfy desired features.
For instance, if the model is producing toxic content, we can apply Reinforcement Learning (RL) techniques to discourage it from producing similar content. 
However, naively applying RL yields \ptmdII{a} model that \ptmdII{can} undergo \emph{catastrophic forgetting} of its original distribution.
For example, it can degenerate into producing a single nonsensical but at least nontoxic sequence. 
%
Although several studies have considered hand-crafting general rewards to ensure desirable features like fluency~\citep{LiuLSNCP16,RL_TambwekarDMMHR19}, coming up with complete or perfect rewards is highly non-trivial~\citep{Wu_googleMT16,VedantamZP15}. This has sparked a wide discussion on the overall effectiveness of RM for some tasks such as machine translation~\citep{Choshen20WeaknessRL,KiegelandK21}.%
\fmdII{The figure is nice, but small print is a bit pale, perhaps make the color darker.}

\paragraph{Reward Maximization with KL-Control}
To tackle the aforementioned issues of ``catastrophic forgetting'', several studies, still under an RM paradigm, have considered incorporating a distributional term inside the reward to be maximized. In particular~\citet{Jaques-2017,KL_jaquesK19} and \citet{Ziegler19} or more recently \citet{ziegler_summarize2021}, \citet{Ouyang}, \citet{bai2022training} and \citet{redteaming} have applied variations of KL-control~\citep{todorov,kappen2012optimal} which adds a penalty term to the reward term so that the resulting policy does not deviate too much from the original one in terms of KL-divergence. The overall objective with the KL-penalty is maximized using an RL algorithm of choice including: PPO~\citep{PPO} as in \cite{Ziegler19} or \cite{bai2022training} or Q-learning~\citep{MnihKSGAWR13} as in~\cite{Jaques-2017}.
Adding this \emph{distributional} KL-penalty to the reward raises some important questions:
%
What effect does it have on the shape of the optimal policy? Does this new objective have any interpretation from a distributional perspective?

\paragraph{Distribution Matching}
A different \ptmd{recent} paradigm for fine-tuning language models to satisfy downstream preferences formulates the problem as Distribution Matching (DM).
This paradigm consists of two steps: first a target distribution incorporating the desired preferences is defined as an Energy-Based Model~\citep{lecun_tutorial_2006}.
%
Then the forward KL divergence is minimized between this target distribution and an auto-regressive policy using a family of algorithms referred to as Distributional Policy Gradients (DPG)~\citep{opt-rl-arxiv-2019,khalifa_2021,korbak2021code,pmlr-v162-korbak22a}. 
This approach capitalizes on the flexibility of EBMs in specifying the target distribution. For example, the EBM can be defined so that it conforms to all downstream preferences while its corresponding normalized distribution has a minimal KL divergence from the original, pre-trained language model, therefore tackling the problem of ``catastrophic forgetting''~\citep{khalifa_2021}.
Interestingly, this DM paradigm can also deal with \emph{distributional} preferences, for instance, for de-biasing language models by specifying that the generated sequences should be gender-balanced, i.e. that 50\% of generations contain female mentions. Such distributional constraints cannot be defined in the RM paradigm where a reward is calculated for a single sequence. 

We can notice the promises and limitations of these two paradigms for fine-tuning language models. RM approaches are equipped with an arsenal of RL algorithms and optimization techniques that \ptmd{can}
be efficient in reward maximization, however they lack the distributional aspect to avoid catastrophic forgetting and impose distributional  preferences over LMs. 
DM approaches are suited to tackle those limitations,
however, the family of DPG algorithms 
\ptmd{currently used}
is not as rich as its RL counterpart.

\ptmdAR{While the connections between these two seemingly distinct paradigms have been noted \citep{opt-rl-arxiv-2019,rl_kl_penalties}, they have not been explored in detail. 
Clarifying such connections might help import ideas from one approach to the other. 
This is our goal in this paper, detailing the nuanced connections and applying them to a case-study in variance reduction.}
Overall, our contributions are the following: 
\begin{itemize}[leftmargin=15px]
    \itemsep0.2em 
    \item  \ptmdAR{We clarify relations between the RM and DM paradigms through a detailed comparison between the family of DPG algorithms and Policy Gradients (Table \ref{tab:comparison_DPGvsPG}), stressing the differences between \emph{parametric} and \emph{non-parametric} rewards that are important in this regard.}  
    \item We introduce an interpretation of KL-control techniques from a distribution matching perspective, placing such techniques at an intermediate place between RM and DM (Theorem \ref{theorem_kl}).
    \item We show how these connections can enable cross-pollination between the two perspectives by applying \emph{baselines} --- a variance reduction technique from RL --- to DPG and derive a particular choice of a baseline (Facts \ref{dpgon_baseline_unbiased} and \ref{dpgoff_baseline_unbiased}). On an array of controllable language generation experiments, 
    we show that adding baselines leads to superior performance on constraint satisfaction (Figure \ref{fig:pointwise-compare-methods-metrics}), stability on small batch sizes, and sample efficiency (Figure \ref{fig:bsz}). 
\end{itemize}

\section{Background}

\smallparagraph{Standard Policy Gradients}
\label{sec:pg}
One popular method for adapting the behaviour of language models to certain
preferences has been that of assigning a ``reward'' score $R(x)$ for sequences $x$ sampled from an autoregressive language model (policy) $\pi_\theta$. 
Then, the simplest policy gradient algorithm in reinforcement learning, namely, REINFORCE \citep{Williams92}, 
aims to find
the policy $\pit(x)$ that maximizes the average reward $\EX{x \sim \pit} R(x)$, and this leads, via the so-called ``log derivative trick'', to a gradient ascent algorithm that iteratively samples $x$ from $\pit$ and update parameters by increments proportional to $R(x)\nabt \log \pit(x)$ via the following identity:
\begin{align}
  \label{eq:REINFORCE}
  \nabt \EX{x \sim \pit} R(x) &= \EX{x \sim \pit} R(x) \nabt \log \pit(x).
\end{align}

\smallparagraph{KL-control}
\label{sec:kl_control}
~\citep{todorov,kappen2012optimal}, was leveraged by \cite{Jaques-2017,KL_jaquesK19} and \cite{Ziegler19} to include a KL penalty term in the reward function to penalize large deviations from the original pretrained model $a(x)$, weighted by a free hyperparameter $\beta$ to control the trade-off between the two goals.
That is, they maximize
the expectation $\E_{x \sim \pit} \Rtz(x)$, where:
\begin{align}
  \Rtz(x)\doteq r(x) - \beta \log \frac{\pit(x)}{a(x)}.
  \label{RTZ_first_mention}
\end{align} 


\smallparagraph{Distributional Policy Gradients}
\label{sec:dpg}
%
%
(DPG)~\citep{opt-rl-arxiv-2019} is 
\ptmd{a recent}
approach used to fit an autoregressive policy $\pit$ to the distribution $p(x)=P(x)/Z$ induced by the EBM $P(x)$, where $Z=\sum_x P(x)$ is the normalization constant (partition function). Given an arbitrary EBM $P(x)$, 
DPG optimizes the loss function $\KL(p, \pit)$ with respect to the parameters $\theta$ of an autoregressive model $\pit$, 
\ptmd{a loss which is minimized for}
$\pit=p$. 
%
The KL-divergence minimization objective leads to a gradient estimate of the form:
\begin{align}
\nabla_\theta \KL(p, \pit) =& - \nabla_\theta \EX{x\sim p} \log \pit(x) \\
=& - \sum_x p(x) \nabla_\theta \log \pit(x) = - \frac{1}{Z}\sum_x P(x) \nabla_\theta \log \pit(x) \\
=& - \frac{1}{Z} \,\,\, \EX{x \sim \pit} \frac{P(x)}{\pit(x)} \nabla_\theta \log \pit(x).\label{eq:dpgon}
\end{align}

\section{Reward Maximization vs Distribution Matching}
In the previous section, we have summarized three approaches that have been suggested for fine-tuning language models. Two of them can be characterized as ``Reward Maximization'' (RM): Standard Policy Gradients (PG) and KL-control. On the other hand, DPG clearly belongs to the realm of ``Distribution Matching'' (DM) as it first defines the target distribution and then optimizes a policy to match it. In the rest of this section, we will explore connections between these two seemingly distinct concepts and, in the following section, we will exploit them to improve DM-based methods.

\subsection{Standard vs. Parametric Rewards}
\allowdisplaybreaks

Let us start with distinguishing between a ``parametric reward'' $\Rtheta$ which depends on $\theta$ and a standard reward $R$, which does not.
If we wished to maximize the expected parametric reward, $\E_{\pit} \Rtheta(x)$, 
we would follow its gradient, leading to the identities:
\begin{align}
    \nabt \E_{x \sim \pit} \Rtheta(x) &= \nabt \sum_x \pit(x) \Rtheta(x) 
    = \sum_x \pit(x) \nabt \Rtheta(x) + \sum_x \Rtheta(x) \nabt \pit(x) \\
    &= \sum_x \pit(x) \nabt \Rtheta(x) + \sum_x \pit(x) \Rtheta(x) \nabt \log \pit(x) \\
    &= \underbrace{\E_{x \sim \pit} \nabt \Rtheta(x)}_{\textrm{RG-term}} + \underbrace{\E_{x \sim \pit} \Rtheta(x) \nabt \log \pit(x)}_{\textrm{PG-term}}. \label{eq:two_terms}
\end{align}

Equation \eqref{eq:two_terms} is the sum of two terms: the first one, the ``RG-term" (Reward Gradient term), involves the gradient of the reward. The second one, the ``PG-term'' (Policy Gradient term), was obtained using the ``log derivative trick'' and involves the gradient of the policy \emph{stricto sensu}. In standard RL, where the reward does \emph{not} depend on $\theta$, the RG-term disappears and the gradient of expected reward consists solely of the PG-term. However, when $\Rtheta$ depends on $\theta$, the gradients are distinct (apart from specific cases where the RG-term evaluates to $0$, as we will see below).

\subsection{KL-control as Distribution Matching}
Adding a KL-penalty term to the reward (as in the case of KL-control)\fmd{Harmonise KL-control vs KL control.} leads to a parametric reward. However, due to the particular form of its objective, the RG-term actually \emph{vanishes},\footnote{This is because $\E_\pit \nabt \Rtz(x) = -\beta\, \E_\pit \nabt \log \pit(x) = 0$, via the 
identity $\E_\pit \nabt \log \pit(x) = \sum_x \pit(x) \nabt \log \pit(x) = \sum_x \nabt \pit(x) = \nabt \sum_x \pit(x) = 0$.}
leaving only the PG-term $\E_{x \sim \pit} \Rtz(x) \nabt \log \pit(x)$ and simplifying the tuning procedure to a standard Policy Gradient.
While this algorithm falls under the RM paradigm, here we argue that is its nature is multifaceted, and explore deeper connections with the DM paradigm. 
More precisely, the maximization of reward with the KL penalty term is equivalent to a distributional matching with an underlying emergent sequential EBM, a remark that already reveals some similarities with DPG.\footnote{The optimal policy $p_z$ is briefly mentioned in \citep{Ziegler19} without reference or derivation. The proof, which 
\ptgk{reveals a connection to the reverse KL divergence from $\pit$},
is ours.}

\begin{theorem}
\label{theorem_kl}
Consider the following EBM:
\begin{equation}
   P_z(x) = a(x)e^{r(x)/\beta} \label{eq:Ziegler_optimal}
\end{equation}
and let $p_z$ be the normalized distribution $p_z(x)  = \frac{1}{Z}\;  P_z(x)$, with $Z=\sum_x P_z(x)$. Then: 
\begin{enumerate}[itemsep=0pt, label=(\roman*)]
    \item $\argmax_\pit \E_{x \sim \pit} \Rtz(x) = \argmin_\pit \KL(\pit, p_z)$;
    \item $\argmax_{\pi \in \mathcal{D}(X)} \E_{x \sim \pi} \Rpiz(x) = p_z$, 
    where $\mathcal{D}(X)$ is the family of all distributions over $X$, and $\Rpiz(x) \doteq r(x) - \beta \log \frac{\pi(x)}{a(x)}$.
\end{enumerate}
\end{theorem}
\vspace{-5px}
\begin{proof}
A simple way to prove  this is to notice that the expectation of the reward $\Rtz$ 
has a monotonically decreasing relationship with the \emph{reverse} KL divergence between $\pit$ and $p_z$:
\begin{align*}
        \KL(\pit, p_z) &= \E_{x \sim \pit} \log \frac{\pit(x)}{p_z(x)}
    = \E_{x \sim \pit} \Big[\log \pit(x) - \log \frac{1}{Z} a(x)e^{r(x)/\beta}\Big] \\
    &= \log Z - \frac{1}{\beta}\, \E_{x \sim \pit} \Big[r(x) -\beta \log \frac{\pit(x)}{a(x)}  \Big] = \log Z - \frac{1}{\beta}\, \E_{x \sim \pit} \Rtz(x), \numberthis \label{eq:klrmax}
\end{align*}
so that the $\argmin_{\pi_\theta} \KL(\pit, p_z)$ coincides with the $\argmax_{\pi_\theta} \E_{x \sim \pit} \Rtz(x)$, proving (i). On the other hand, $\argmin_{\pi \in \mathcal{D}(X)} \KL(\pi, p_z)$, which also corresponds to $\argmax_{\pi \in \mathcal{D}(X)} \E_{x \sim \pi} \Rpiz$ because of (i) applied to a family $\pi_{\theta'}$ covering $\mathcal{D}(X)$ in full, is just $p_z$, concluding the proof.
\end{proof}
Overall, we can conclude that the addition of the distributional term (KL-penalty) to the reward does indeed provide a DM interpretation, namely in terms of minimizing the reverse KL divergence with an emergent underlying distribution $p_{z}(x)$. We note that $p_{z}(x)$ 
\ptmd{does not correspond to a free and explicit choice of EBM} 
(e.g. one that balances the gender and topic distributions of a language model). Instead equation~\eqref{eq:Ziegler_optimal} 
appears in a restrictive format, which is
implicitly defined by the reward $\Rtz$, along with a $\beta$ hyperparameter without a clear meaning. By contrast, the DPG algorithms are designed to perform DM on any EBM specification, corresponding to an explicit distributional objective.

\subsection{Similarities and Differences between DPG and Policy Gradients}
In the previous subsection, we have connected KL-control, a method designed under a RM paradigm, to DM. Now, we turn to the converse question of whether DPG, a DM method, can be connected to RM.
We begin by noting that after defining $\Rt = \frac{P(x)}{\pit(x)}$, the DPG gradient $\EX{x \sim \pit} \frac{P(x)}{\pit(x)} \nabla_\theta \log \pit(x)$ acquires the format of the PG-term $\EX{\pit} \Rt \nabla_\theta \log \pit(x)$.

However, the DM objective of DPG \emph{cannot} be considered as maximizing the average ``reward'' $R_\theta(x) = \frac{P(x)}{\pit(x)}$, as this would require adding also the RG-term $\E_{\pit}\nabla_\theta \frac{P(x)}{\pit(x)}$ into the gradient, which in this case does not vanish. 

Nonetheless, the analogy 
\ptmd{behind}
this gradient term is more fruitful than it first appears.
As a matter of fact, DPG gradient estimates suffer from the same high-variance problems as with standard PG. While the objective of DPG (distribution matching) is different from that of Policy Gradients (reward maximization), DPG also needs to estimate the PG-term $\E_\pit \Rtheta(x) \nabt \log \pit(x)$ at a \emph{given} value of $\theta$, using a batch of samples $x$. 
 For such a \emph{fixed} $\theta$, we can define provisionally set $R(x)\doteq \Rtheta$ and the problem of gradient estimation \emph{for this fixed $\theta$} is 
 \ptmd{identical}
 to the estimation $\EX{x\sim \pit} R(x) \nabla_\theta \log \pit(x)$ based on a set of samples $x$ in standard RL.
Therefore, techniques that have been developed to reduce the variance of the gradients estimates in RL can be ported to DPG insofar as we are computing the gradient estimates \emph{at a given $\theta$}. %
In Section \ref{section:method}, we show how one can import one such variance reduction technique to the DPG: baselines.






\section{A Case Study on Variance Reduction}\label{section:method}

\begin{table*}[t]
    \centering
    \small
    \vspace{-0.5cm}
\begin{tabular}{lll}
\toprule
 & \textbf{Policy Gradients} & \textbf{\DPG}\\
\midrule
 \textbf{Reward} & 
 $R(x)$   & 
 $R_\theta(x) =\frac{P(x)}{\pi_\theta(x)}$ \\
 \addlinespace
 \textbf{ $\nabla_\theta$} & 
{\small $\E_{x \sim \pit} R(x) \nabt \log \pit(x)$} & 
{\small $\E_{x \sim \pit} \frac{P(x)}{\pi_\theta(x)} \nabt \log \pit(x)$} \\
 \addlinespace
 \textbf{Baseline}& 
 $\E_{x \sim\pit} R(x)$  & 
 $Z$   \\
\addlinespace
 \textbf{ $\nabla_\theta$ with Baseline } & 
 {\small $\E_{x \sim \pit} \Big[R(x) - \E_{x \sim\pit} R(x) \Big] \nabt \log \pit(x)$} & 
{\small $\E_{x \sim \pit} \Big[ \frac{P(x)}{\pi_\theta(x)} - Z \Big] \nabt \log \pit(x)$} \\
\bottomrule
\end{tabular}
    \caption{\small{A comparison between Policy Gradients~\citep{sutton_policy_gradients} and Distributional Policy Gradients~\citep{opt-rl-arxiv-2019} forms of Reward, Baseline, and Gradient of the loss function (the PG-term) before ($\nabla_\theta$) and after ($\nabla_\theta$ with Baseline) including a baseline for variance reduction .}}
    \label{tab:comparison_DPGvsPG}
\end{table*}

Baselines are a standard variance reduction technique in the context of Policy Gradients \citep{Sutton2018}. The idea is to subtract from the reward $R(x)$ a value $B$ 
that does not introduce bias to the gradients but may change variance. After the introduction of baseline, equation \eqref{eq:REINFORCE} then takes the following form:
\begin{equation}
\label{pg_baseline}   
\nabt \EX{\pit} R(x) = \EX{\pit} [R(x)-B]\, \nabt \log \pit(x).
\end{equation}

\begin{wrapfigure}{r}{0.4\textwidth}
    \vspace{-0.8cm}
    \centering
    \includegraphics[width=0.4\textwidth]{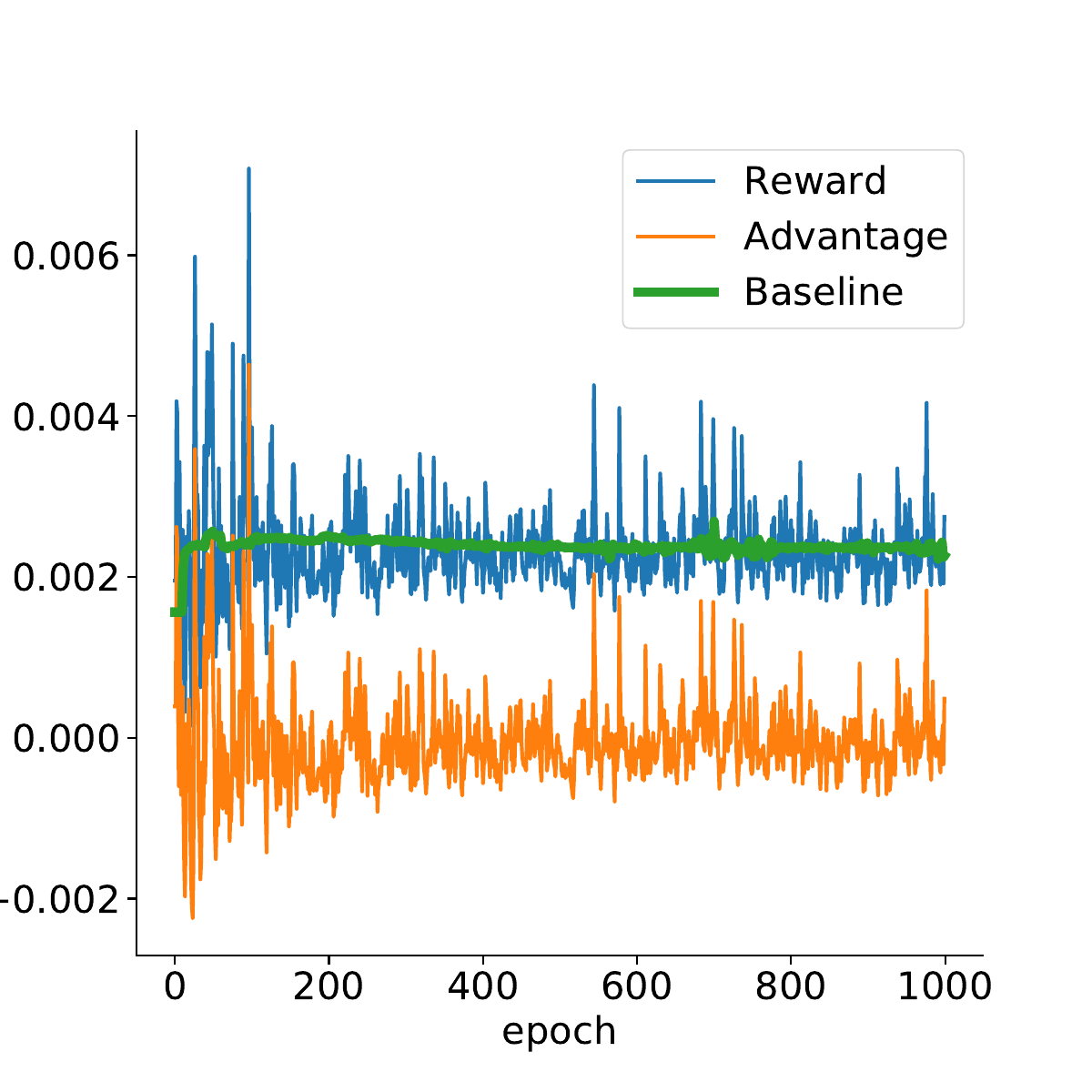} 
    \caption{
    \small{Values of reward, advantage and the baseline  for first 1000 epochs of a pointwise constraint experiment.}
    }
    \label{fig:advantage-illustration}
\end{wrapfigure}

In standard RL, the simplest form of baseline $B$ is just the average of the rewards for the policy:\footnote{While this baseline is not optimal (proof Appendix \ref{appendix:optimal-baseline}), it is widely used in practice.}
\begin{equation}
    \label{eq:B_rl}
    B^{\text{RL}} = \E_{x \sim \pi_\theta} R(x).
\end{equation}

Following the same methodology of taking the baseline to be the expectation of the reward term, we can obtain a remarkably simple form of a baseline for DPG:
\footnote{In the scope of this paper, our focus is on  importing to DPG simple constant baselines. The advantage is that this is a technique that is not impacted by the fact that $R_\theta$ depends on $\theta$: it can be applied ``$\theta$-locally'' to provide a more accurate estimate of $\E_{x \sim \pit} R_\theta(x) \nabla_{\theta} \log \pit(x)$ for a \emph{fixed} $\theta$, irrespective of the values of $R_{\theta'}$ elsewhere, while variance reduction techniques that involve several $\theta's$ simultaneously raise additional challenges for parametric rewards.
}
\begin{equation}
 \label{eq:B}
\begin{aligned}
    B = \E_{x\sim \pit} \frac{P(x)}{\pit(x)} = \sum_x \pit(x) \frac{P(x)}{\pit(x)} = \sum_x P(x) = Z.
\end{aligned}
\end{equation}

\begin{fact}
\label{dpgon_baseline_unbiased}
Subtracting 
$B$ 
from $R_\theta(x)$ does not introduce bias into DPG gradient estimates.%

\end{fact}
\vspace{-10px}
\begin{proof}
Let us rewrite the DPG gradient in \eqref{eq:dpgon} with the added baseline $B=Z$:
\begin{equation}
\begin{aligned}
   \E_{x \sim \pit} \Big[ R_\theta(x) - Z \Big] \nabla_{\theta} \log \pit(x)
    &=\E_{x \sim \pit} R_\theta(x) \nabla_{\theta} \log \pit(x) 
    -Z \,\E_{x \sim \pit} \, \nabla_{\theta} \log \pit(x) \\
    &=\E_{x \sim \pit} R_\theta(x) \nabla_{\theta} \log \pit(x) 
    - Z \Big[\sum_{x} \nabla_{\theta} \pit(x) \Big]
\end{aligned}
\end{equation}
Here, the second term does not introduce bias because $Z \Big[\sum_x \nabt \pit(x) \Big]= 0$, leaving us with the exact same form of gradient as in the original DPG algorithm.
\end{proof}

\begin{wrapfigure}{r}{0.53\textwidth}
\begin{minipage}{0.53\textwidth}
\vspace{-13px}
\begin{algorithm}[H]
\caption{KL-Adaptive DPG  \textcolor{blue}{with baseline} \label{al:KL-adaptive-DPG-baseline}}
\begin{small}
\begin{algorithmic}[1]
\Require $P$, initial generative model $a$
\State $\pi_\theta \gets a$, $q \gets a$
\For{each iteration}
\For{each episode}
    \State sample $x$ from $q(\cdot)$
    \State $\theta \gets \theta + \alpha^{(\theta)} \textcolor{blue}{\Big[ \textcolor{black}{\frac{P(x)}{q(x)}} - Z\frac{\pit(x)}{q(x)} \Big]} \ \nabla_\theta \log \pi_\theta(x)$ 
\EndFor
\If{ $\KL(p||\pi_\theta) <  \KL(p||q)$} 
    \State $q \gets \pi_\theta$
\EndIf
\EndFor
\Ensure $\pi_\theta$
\end{algorithmic}
\end{small}
\end{algorithm}
\vspace{-20px}
\end{minipage}
\end{wrapfigure}
Note that since $B^{\text{RL}}$ depends on $\theta$, it has to be be re-estimated after each gradient update. On the other hand, $B$ does \emph{not} depend on $\theta$, which is an advantage because $B$ could be now estimated by averaging over samples from \emph{all} the different $\theta$'s without introducing bias, leading to a more accurate estimation. See Table \ref{tab:comparison_DPGvsPG} for a comparison of these two forms of baselines.\fmd{(4-10) About Table 1: in the caption you speak about $\nabt$ in the left column referring to the gradient of the loss function. I think this might be misleading (several forms of loss functions in our paper), what about replacing $\nabt$ in the first column with PG-term, or at least using PG-term terminology in the caption?}

The off-policy DPG version introduced in~\citep{opt-rl-arxiv-2019} and its KL-adaptive variant~\citep{khalifa_2021} sample a proposal distribution $q$ instead of the policy $\pit$. Then, the baseline takes the form
\begin{equation}
    \Boff = Z\frac{\pit(x)}{q(x)},
\end{equation}
where the $\frac{\pit(x)}{q(x)}$ term is an importance weight correcting for the bias introduced by sampling from~$q$. Similarly to the DPG case, we can prove the following (see Appendix~\ref{appendix:baselines}):
\begin{fact}
\label{dpgoff_baseline_unbiased}
Subtracting $\Boff$ from $R_\theta(x)$ does not bias the off-policy DPG gradient estimates.
\end{fact}

In practice, as shown on Figure~\ref{fig:advantage-illustration}, adding a baseline to KL-adaptive DPG  (\Algref{al:KL-adaptive-DPG-baseline}) centers the advantage 
\ptmd{(defined as $A \doteq  \frac{P(x)}{q(x)} - Z\frac{\pit(x)}{q(x)}$)}
around 0 leading to better performance on: convergence (section \ref{subsec:general_results}), stability on small batch sizes (section \ref{subsec:bsz_exps}), and variance reduction (section \ref{subsec:effectOnVarReduc}).

\subsection{Generation with Distributional Control}


%
We investigate the benefits of adding a baseline to the DPG algorithm, on the Generation with Distributional Control (GDC)~\citep{khalifa_2021} framework. %
GDC makes use of DPG to control the properties of pre-trained language models to satisfy certain constraints. 
In our experiments, follow target distribution form of  \citet{A-parshakova-etal-2019-global}, \citet{khalifa_2021} and \citet{pmlr-v162-korbak22a}, in which the EBM $P(x)$ is defined  so that its  normalized variant $p(x)$
matches a set of desired moments constraints on given features $\phi_i(x)$, while having a minimal KL divergence $\KL(p,a)$ from an original pretrained language model $a$, to avoid catastrophic forgetting. 

These constraints are expressed as conditions $\bar{\mu}_i = \EX{x \sim p} \phi_i(x)$, for $i \in \{1,\dots,n\}$, by which the moments (expectations) under the distribution $p$ of each feature $\phi_i(x)$ are required to take certain desired values $\bar{\mu}_i$. For instance, let $\phi_1(x) = 1$ iff the topic of $x$ is science and $\phi_2(x) = 1$ iff $x$ mentions a female person, then imposing moments $\bar{\mu}_1 = 1$ and $\bar{\mu}_2 = 0.5$ constrains the language model $p$ to only generate sequences about science, half of which mention females. $P(x)$ is uniquely determined by the following  form:\footnote{For a more precise formulation of this EBM, see \citep{khalifa_2021}.}
\begin{equation} 
    P(x) = a(x)e^{\sum_{i=1}^n \lambda_i \phi_i(x)}, \label{eq:target-EBM}
\end{equation}
where $\lambda_i$ terms control the moments $\mu_i$ of the associated features, which can be estimated through self-normalized importance sampling \citep{owen_chapter_importance_sampling_2013}; and then, to make the moments match the desired values, the $\lambda_i$ terms can be optimized through SGD~\citep{A-parshakova-etal-2019-global}.


%
%
%
%
\subsection{Experimental setup}\label{subsec:exp_setup}

We evaluate our method on an array of 10 controlled text generation tasks. For each, given a pre-trained language model $a(x)$, and a set of constraints, the objective of each fine-tuning method is to obtain a fine-tuned language model $\pit$ that satisfies the imposed constraints while deviating as minimally as possible from the original language model $a(x)$. 

Constraints are defined as a set of binary features $\{\phi_i\}$ and their corresponding desired percentages (moments) $\{\bar{\mu}_i\}$ within the generations of the target language model.
%
Based on the value of the moment constraints these 10 tasks are divided into 6 tasks of pointwise constraints (for which  $\bar{\mu}_i = 1$), 2 tasks of distributional constraints ($0 < \bar{\mu}_i < 1$) and 2 tasks of mixed type constraints (hybrid):
\vspace{-0.1cm}
\begin{enumerate}[label=(\alph*)]
    \itemsep0em 
    \item Single-word constraints, where $\phi(x) = 1$ iff the a given word appears in the sequence $x$. We experiment with frequent words (task 1: ``amazing'', original frequency: $10^{-4}$) and (task 2: ``WikiLeaks'', original frequency: $10^{-5}$) rare words,
    \item Wordlist constraints, where $\phi(x) = 1$ iff $x$ contains at least one word from a given list. We consider lists of word associated with politics (task 3) and science (task 4) published by \citet{plug_and_play_20},
    \item Sentiment classifier constraints, where $\phi(x) = 1$ if $x$ is classified as positive (task 5), or negative (task 6) by a pre-trained classifier published by \citet{plug_and_play_20}.
    \setcounter{enumi}{3}
    \item A single distributional constraint where $\phi(x) = 1$ iff $x$ contains a female figure mention, and $\bar{\mu} = 0.5$ (task 8),
   \item A set of four distributional constraints: $\phi_i(x) = 1$ iff $x$ contains at least one of the words in the ``science", ``art", ``sports" and ``business" wordlists (compiled by \citet{plug_and_play_20}), respectively. For each $i$, $\bar{\mu}_i = 0.25$ (task 8),
    \item Hybrid constraints where $\phi_1(x) = 1$ iff $x$ contains more female than male pronouns, $\bar{\mu}_1 = 0.5$ and $\phi_2(x) = 1$ iff $x$ contains at least one of the words from the ``sports" wordlist (task 9) or ``politics'' wordlist, $\bar{\mu}_2(x) = 1$ (task 10).
\end{enumerate}
\vspace{-0.2cm}
\paragraph{Methods}
We modify the GDC framework~\cite{khalifa_2021}, namely its KL-DPG algorithm, to include a baseline as shown in \Algref{al:KL-adaptive-DPG-baseline}. We refer to this method as \textbf{GDC\texttt{++}}. 
In addition to comparing \textbf{GDC\texttt{++}} with \textbf{GDC}
we compare with two reward maximization baselines: \textbf{Reinforce} \citep{Williams92Reinforce} and \textbf{Ziegler} \citep{Ziegler19}. Reinforce tries to maximize the expected reward $\E_{x \sim \pit} R(x)$, where $R(x) =1$ if and only if the pointwise constraints are met. Ziegler instantiates the KL-control approach: its objective includes a KL penalty term for departures from $a$.
Following~\citep{khalifa_2021}, for hybrid and distributional constraints (tasks 8-10) we compare only GDC and GDC\texttt{++} because the RM objective of Ziegler and Reinforce is not equipped to handle them. 

\vspace{-5px}
\paragraph{Metrics} 
We report the following metrics at each validation step over batches of samples from $\pit$:
\vspace{-0.3cm}
\begin{enumerate}
    \itemsep0em 
    \item $\E_{x \sim \pit} \phi_i(x)$, measuring the ability to reach the target moment of the $i$-th feature. 
    \item $\KL(p, \pit)$, the forward KL divergence from the optimal target distribution $p$,\footnote{See Appendix \ref{detailed-metrics} for a detailed description of how $\KL(p, \pit)$ is computed.}
    \item $\KL(\pit, a)$, the reverse KL divergence from the original pretrained language model $a$. 
    \item Distinct-n score, a measure of text diversity in terms of the frequency of repetitions within a single sample $x$, proposed by \citep{li-etal-2016-diversity}. 
    \item Self-BLEU-n, a measure of text diversity on a distributional level \emph{across} samples proposed by \citep{texygen-ZhuLZGZWY18}, ensuring that policies don't converge into limited number of sequences that satisfy the imposed constraints~\cite{GAN_short}.
\end{enumerate}
\vspace{-10px}
\paragraph{Training details}
For tasks 1-6, we use a pre-trained GPT-2 small with 117M parameters \citep{radford2019language} as the original language model $a$. For tasks 7-10, $a$ is the same pre-trained model additionally fine-tuned on the WikiBio~\citep{DBLP:conf/emnlp/LebretGA16} dataset. See Appendix \ref{appendix:Hyperparameters} for more details. The code for all the experiments presented in the paper will be available at \href{https://github.com/naver/gdc}{github.com/naver/gdc}.

\subsection{Results}
\label{subsec:general_results}
\begin{figure*}[t]
    \centering
    \vspace{-0.5cm}
    \includegraphics[width=\linewidth]{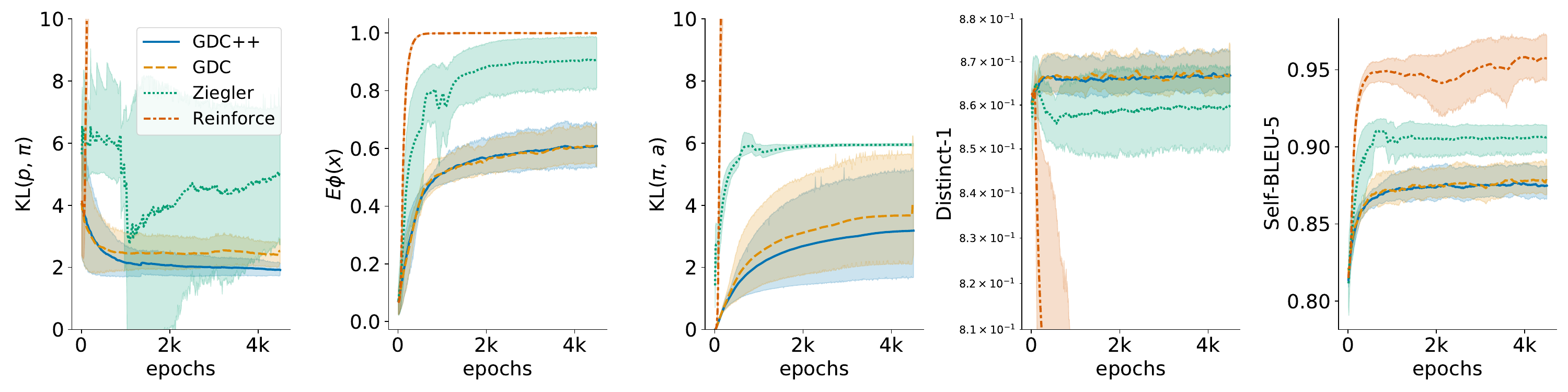} 
    \caption{\small{Evaluation metrics:  $\KL(p, \pi_{\theta})$ ($\downarrow$ better), $\mathbb{E}_{\pit} \phi(x)$ ($\uparrow$ better), $\KL(\pi_{\theta}, a)$ ($\downarrow$ better), Self-BLEU-5 ($\downarrow$ better), and Distinct-1 ($\uparrow$ better) aggregated over 6 pointwise constraints experiments (tasks 1-6) for policies obtained from GDC\texttt{++}, GDC, Ziegler and Reinforce. See Figure~\ref{fig:distributional-compare-methods-metrics} for aggregated distributional constraints experiments. In the Appendix Figures \ref{fig:pointwise-compare-methods-split1}-\ref{fig:distributional-compare-methods-split} and Table~\ref{tab:all_experiments_results} contain individual view and final results of each run.}}
    \label{fig:pointwise-compare-methods-metrics}
\end{figure*}

We present the evolution of our metrics through training epochs in Figure \ref{fig:pointwise-compare-methods-metrics} (aggregated over tasks 1-6) and Figure \ref{fig:distributional-compare-methods-metrics} in the Appendix (aggregated over tasks 7-10). Results for each task are presented separately on Figures \ref{fig:pointwise-compare-methods-split1}-\ref{fig:distributional-compare-methods-split} in the Appendix.

Consistent with prior work~\citep{khalifa_2021,pmlr-v162-korbak22a}, we observe that Reinforce is able to quickly achieve high levels of constraint satisfaction, but at the cost of large deviations from $a$, which translates into significantly decreased diversity of generated samples (in terms of Self-BLEU-5 and Distinct-1). The KL penalty term in Ziegler imposes an upper bound on deviation from $a$ but the deviation is still significant enough to result in a drop in diversity. Moreover, we have observed Ziegler's objective to result in very unstable training. 

\GDC and \GDCplus are the only fine-tuning methods that address constraint satisfaction based on a clear formal objective, i.e. reducing the divergence from $p$. The approach translates into significantly smaller deviations from $a$ and maintaining diversity within and across samples. The addition of a baseline indeed reduces the variance. We analyze that extensively in Appendix \ref{subsec:effectOnVarReduc} while here focusing on the downstream effects of variance reduction. One is that $\pit$ is now able to compound staying closer to $p$ and $a$ \emph{at the same time}, while achieving slightly better constraint satisfaction. We have also observed that baseline stabilizes training, leading to smoother curves.\footnote{The interested reader can compare the large fluctuations of the Ziegler objective to more stable training curves of \hbox{\GDC,} and even more of \GDCplus, in the disaggregated curves in Figures \ref{fig:pointwise-compare-methods-split1}-\ref{fig:distributional-compare-methods-split} of the Appendix.} 
\begin{figure*}[t]
    \centering
    \begin{subfigure}[t]{.51\textwidth}
    \vskip 0pt
    \centering
    \includegraphics[width=\linewidth]{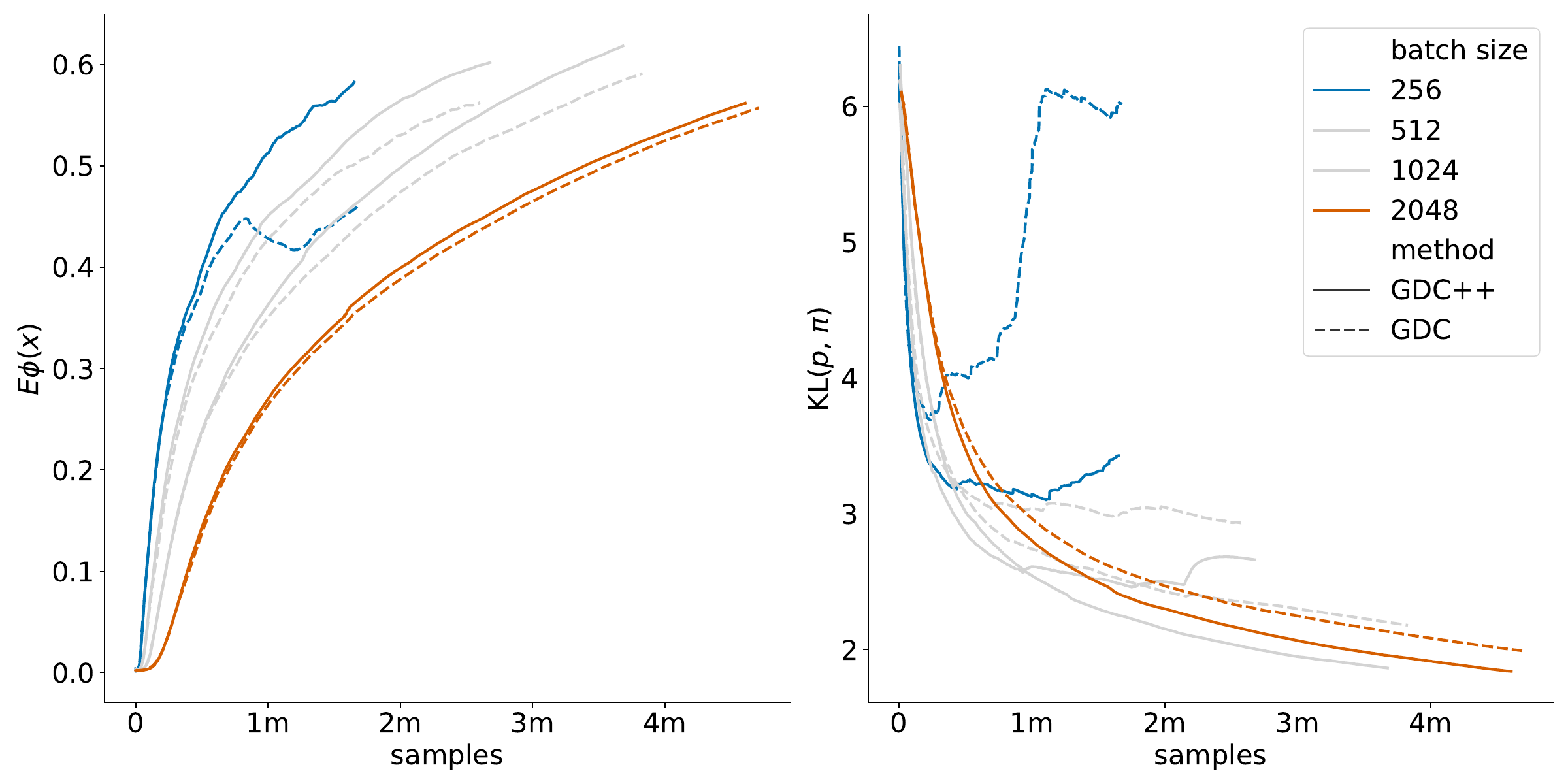} 
    \caption{\small{Task 1: a pointwise constraint}}
    \label{fig:gdc_batch_size_point}
    \end{subfigure}%
    \begin{subfigure}[t]{.51\textwidth}
    \vskip 0pt
    \centering
    \includegraphics[width=\linewidth]{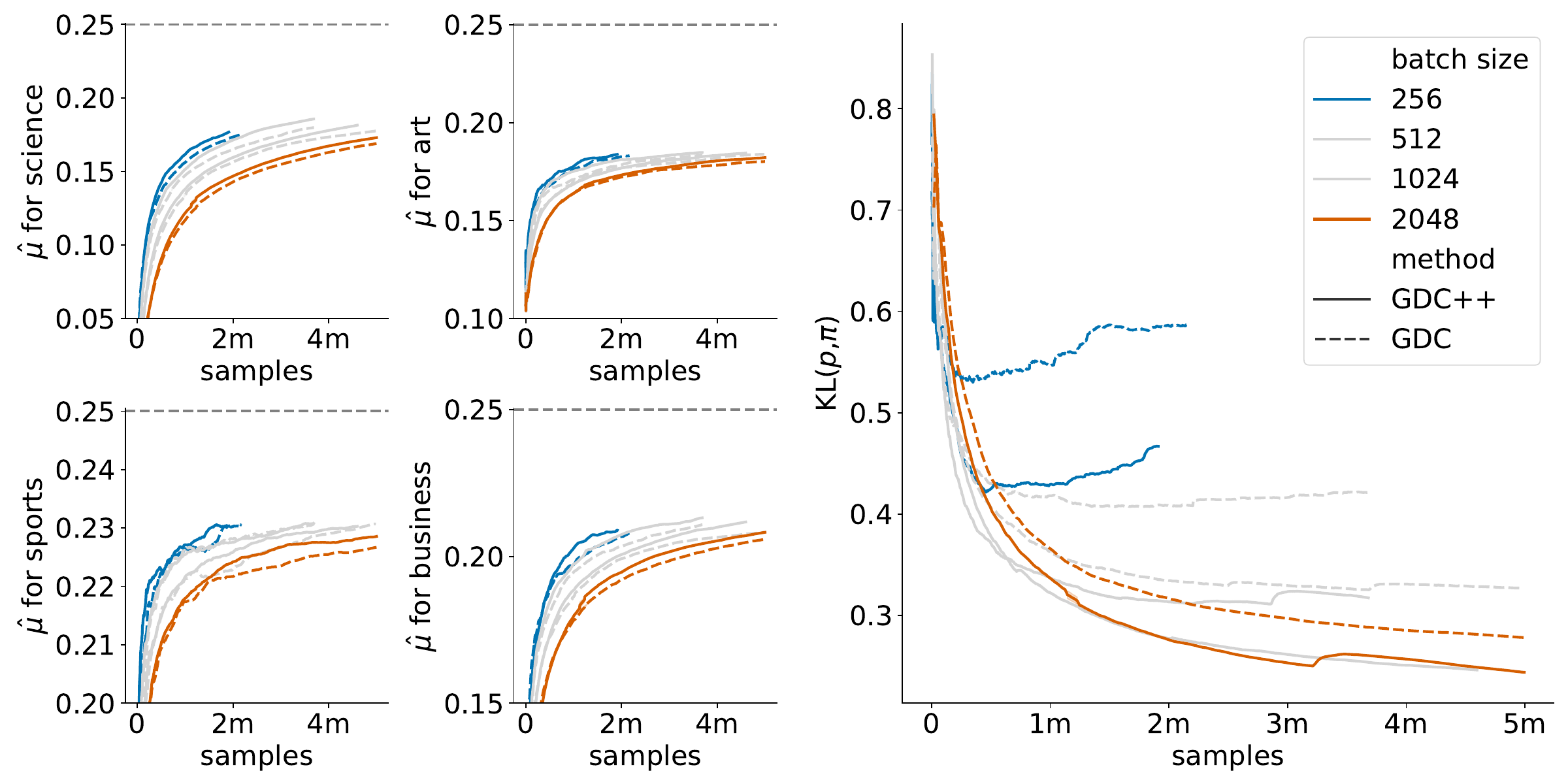} 
    \caption{\small{Task 8: a set of distributional constraints; $\bar{\mu}_i = 0.25$}}
    \label{fig:batch_size_dist}
    \end{subfigure}

    \caption{\small{
    $\mathbb{E}_{\pit} \phi(x)$ or $\hat{\mu}$ per constraint ($\uparrow$ better) and $\KL(p, \pi_{\theta})$ ($\downarrow$ better) as a function of the number of samples reported for task 1 (a) and task 8 (b). We report the number of samples (i.e. the number of epochs times the batch size) for a fair comparison of convergence speed. \emph{\GDCplus is consistently superior across all batch sizes in terms of convergence and constraint satisfaction}. The effect is more conspicuous with small batch sizes. 
    Batch sizes 512 and 2014 are greyed out for clarity. 
    }}
    \label{fig:bsz}
\end{figure*}

\vspace{-5px}
\subsection{The effect of baseline across batch sizes}
\label{subsec:bsz_exps}
We expect that reducing gradient estimates variance can allow to train with lower batch sizes, performing gradient updates on estimates based on smaller batch sizes can increase the sample efficiency. 
To test this, we rerun tasks 1 (a pointwise constraint on the word ``amazing") and 8 ( distributional constraints on topics) with four batch sizes (256, 512, 1024, 2048). 
Figures \ref{fig:gdc_batch_size_point} and \ref{fig:batch_size_dist} show the benefits of adding a baseline --- higher constraint satisfaction, lower divergence from $p$, more stable training --- and is especially evident with lower batch sizes. 
For instance, with batch size 256, \GDCplus obtains a significantly higher constraint satisfaction rate and lower divergence from $p$.

Furthermore, stable training with smaller batch sizes translates into better sample efficiency. For instance, in task 1 (Figure \ref{fig:gdc_batch_size_point}), \GDCplus with batch size 256 needs 1M samples to achieve $\E_{x \sim \pit} \phi(x) = 0.5$ while \GDCplus with batch size 2048 needs 4M. In contrast, \GDC with batch size 256 does not achieve $\E_{x \sim \pit} \phi(x) = 0.5$ at all, confirming the importance of adding the baseline.


\begin{wrapfigure}{r}{0.47\textwidth}
    \centering
    \includegraphics[width=1\linewidth]{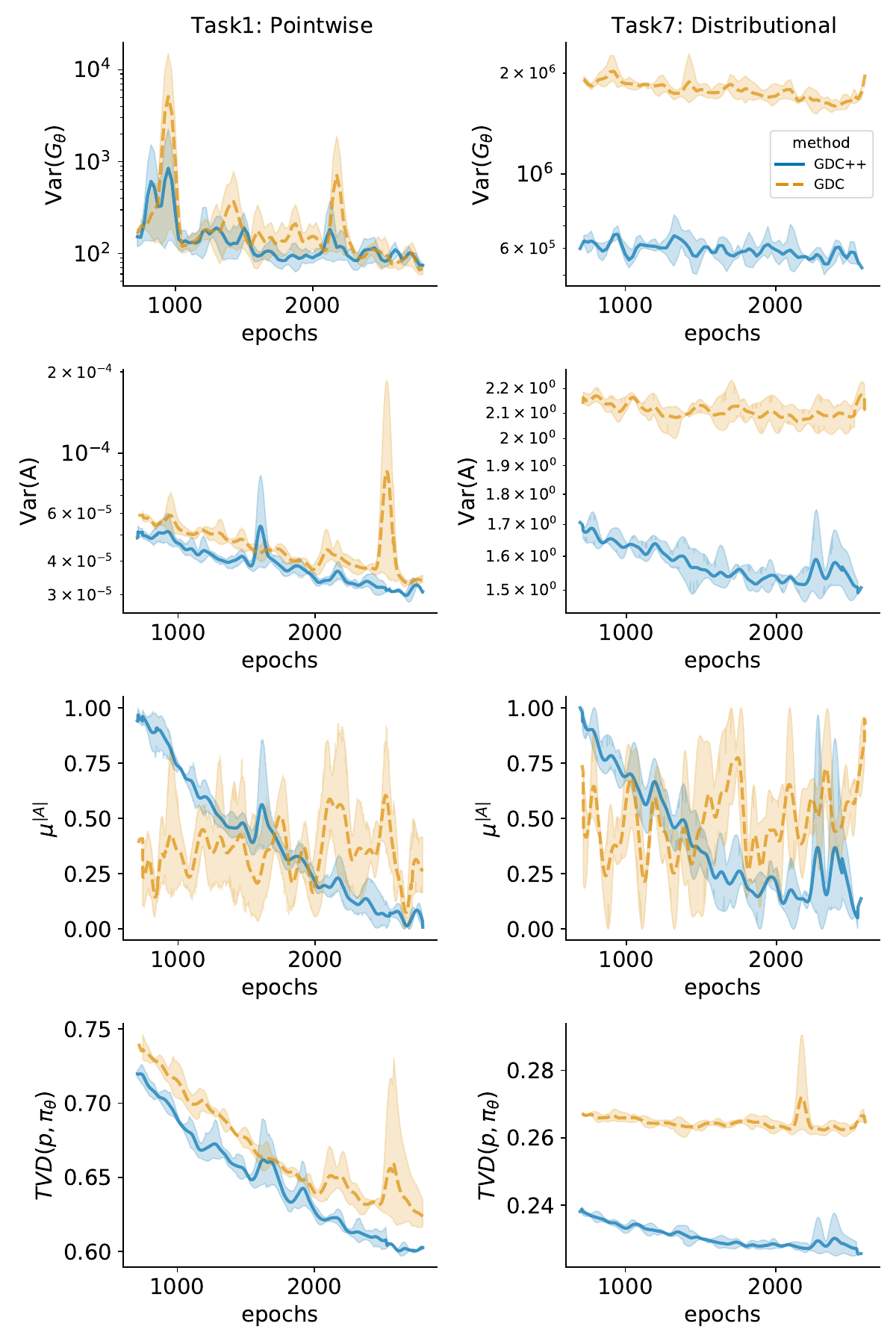} 
    \caption{\small{
    Comparison between \GDC and \GDCplus using a set of Variance diagnosis metrics on pointwise and distributional constraints experiments.
    }}
    \label{fig:grad-main}
    \vspace{-2cm}
\end{wrapfigure}

\subsection{Empirical Evaluation of Variance Reduction}
\label{subsec:effectOnVarReduc}
Next, we evaluate empirically the effect of the baseline for variance reduction. We select two tasks: task $1$ (a pointwise constraint) and task $7$ (distributional constraints) described in Section~\ref{subsec:exp_setup}, each with 3 different seeds, while monitoring the following variance measures: 
\paragraph{Gradient Variance}
The gradient estimate is defined as:
$\grad(x) \doteq A(x) \nabla_{\theta} \log \pit(x)$, where $\grad(x) \in \mathbb{R}^{|\theta|}$ is an unbiased estimate of the gradient of the forward KL loss $\nabla_\theta \KL(p,\pi_\theta)$ with respect to the parameters $\theta$. We then have, with $\mgrad \doteq \mathbb{E}_{x \sim q} \grad(x)$: 
\begin{align}
\vargrad &\doteq \mathbb{E}_{x \sim q} \left\|\grad(x)-\mgrad \right\|_{2}^{2} \\ &=\mathbb{E}_{x \sim q}||\grad(x)||_{2}^{2}- ||\mgrad||_{2}^{2}.
\end{align}
%
\
\paragraph{Variance of the advantage} is defined by:
\begin{align}
\begin{split}
\varAdv &\doteq \mathbb{E}_{x \sim q} \left\|\Adv(x)-\mAdv \right\|_{2}^{2} \\
\end{split}    
\end{align}
where, $\mAdv \equiv \mathbb{E}_{x \sim q} \; \Adv(x)$ is the mean of the advantage, which we showed above to be null after the addition of the baseline. 

\paragraph{Expected absolute value of the advantage}
This metric is defined as: \begin{align}
\begin{split}
\mAbsAdv &\doteq \mathbb{E}_{x \sim q} \;  |\Adv(x)| .
\label{eq:absadv}
\end{split}
\end{align}
It directly provides a standard measure of distributional discrepancy between $p$ and $\pit$, in terms of TVD (Total Variation Distance).
We have: \begin{equation}
\begin{split}
   \E_{x \sim q} \left| \frac{p(x)}{q(x)} - \frac{\pit(x)}{q(x)} \right|   &=
   2\,\TVD(p, \pit).
\end{split}
\end{equation}
%
\paragraph{Results} Figure \ref{fig:grad-main} shows that \GDCplus obtains lower variance in the gradient estimates $\vargrad$ and the variance of the advantage $\varAdv$ in both pointwise and distributional experiments compared to its non-baseline counterpart GDC.
We further observe a decreasing trend in the mean absolute value of the advantage $\mAbsAdv$ which is correlated with a decreasing trend in the TVD distance between the trained policy $\pit$ and the optimal distribution $p$.
Overall, these results shows that adding a baseline to DPG reduces the variance during training and yields better convergence towards the optimal distribution $p$.

\section{Related work}

The idea of posing control problems as distribution matching \ptmdII{has} resurfaced numerous times in the RL literature \citep{kappen2012optimal,friston2010action,levine2018reinforcement,hafner2020action,BUCKLEY201755}. KL-control can be seen as \ptmdII{a} generalisation of maximum entropy RL (MaxEnt RL) \citep{pmlr-v70-haarnoja17a,sac} to informed priors. If in \eqref{RTZ_first_mention} we chose $a(x)$ to be a uniform distribution \ptmdII{(assuming right now finiteness of $X$)}
\fmdII{Over an infinite $X$, one cannot introduce a uniform distribution!} 
instead of a pretrained LM distribution, then the KL penalty $\KL(\pit, a)$ would reduce to an entropy bonus. Both KL-control and MaxEnt RL can be derived from a general framework of control-as-inference \citep{levine2018reinforcement} which poses control as minimising KL from a certain target distribution. However, most practical algorithms in the MaxEnt RL family minimise KL from a target policy which changes throughout training; in contrast, DPG’s target distribution $p$ and KL-control implicit target distribution $p_z$ are defined at trajectory level and fixed throughout training. 

Perhaps the closest method to KL-control and DPG in the larger family of inference-based RL \citep{coadaptation} is AWR \citep{awr} which minimises the \emph{forward} KL from an off-policy target distribution. Yet another approach with apparent similarity to KL-control and DPG is state marginal matching (SMM) \citep{maxentexplor,smm2019}. SMM poses exploration as learning a policy that induces a state marginal distribution that matches a target state distribution. While SMM's target distribution is fixed, it is defined for \ptmdII{individual} states,
\ptmdII{while in the controllable language generation tasks we consider, the target distribution is defined over a complete trajectory considered as a unit.}
See Appendix \ref{appendix:related} for an extended discussion of related work.

\section{Conclusion}\label{section:discussion}
Fine-tuning large language models has become an active area of research, due to its importance in adapting large language models to satisfy task-level preferences, or in combating their social risks such as “distributional” stereotyping~\citep{ethical_lm_deepmind, ethical_lm_deepmind2}.
\footnote{See Appendix \ref{section:Broader} for a discussion of broader impacts of large language models and controllable language generation.}
In this paper, we analyzed in depth the nuanced relation between two popular fine-tuning paradigms: RM and DM. We demonstrated that KL-control can be seen as \ptmd{a form of} DM and showed that while 
\ptmd{DPG and PG have different goals,
some similarities}
(similar forms of gradient estimates despite different objectives) can be exploited. We used these insights to inform an extension of DPG, consisting in adding a baseline to reduce the variance of gradient estimates.

The connections we established suggest that despite fundamental differences between DPG and RL, some of the theoretical results and algorithmic techniques from RL can be adapted to a DM framework without losing their 
\ptmd{formal}
guarantees. In this paper, we focus on variance reduction using baselines, but the space of possible enhancements is vast. Promising candidates include further reducing the variance using a learned value function \citep{conda_actor} and preventing detrimentally large policy updates by maintaining a trust region in the policy space -- akin to techniques such as TRPO \citep{schulman2015trust} and PPO \citep{schulman2017proximal}. \ptmd{Another future direction} could consist in analyzing the relation between explicit EBMs in DPG and implicit EBMs arising in KL-control and characterizing the space of EBMs 
\ptmd{that could be reached through KL-control.}

\clearpage

\bibliography{references}
\bibliographystyle{iclr2022_conference}
\newpage
\section*{Checklist}

\begin{enumerate}

\item For all authors...
\begin{enumerate}
  \item Do the main claims made in the abstract and introduction accurately reflect the paper's contributions and scope?
    \answerYes{}
  \item Did you describe the limitations of your work?
    \answerYes{In Section \ref{section:discussion}}.
  \item Did you discuss any potential negative societal impacts of your work?
    \answerYes{See Appendix \ref{section:Broader}.}
  \item Have you read the ethics review guidelines and ensured that your paper conforms to them?
    \answerYes{}
\end{enumerate}

\item If you are including theoretical results...
\begin{enumerate}
  \item Did you state the full set of assumptions of all theoretical results?
    \answerYes{}
        \item Did you include complete proofs of all theoretical results?
    \answerYes{In Appendix \ref{appendix:baselines} we present proofs of all mathematical facts referred to in the paper.}
\end{enumerate}

\item If you ran experiments...
\begin{enumerate}
  \item Did you include the code, data, and instructions needed to reproduce the main experimental results (either in the supplemental material or as a URL)?
    \answerYes{The code is included as supplementary material available to the reviewers and area chairs and will be made publicly available alongside the camera ready version of the paper.} 
  \item Did you specify all the training details (e.g., data splits, hyperparameters, how they were chosen)?
    \answerYes{In Appendix \ref{appendix:Hyperparameters} we provide the hyperparameters used throughout our experiments and report our hardware configuration. In Appendix \ref{detailed-metrics}, we describe in detail how $\KL(p, \pit)$ and $\TVD(p, \pit)$ were estimated and provide an extended pseudocode for our training loop in Algorithm~\ref{appendix:al:DPG}. }
        \item Did you report error bars (e.g., with respect to the random seed after running experiments multiple times)?
    \answerNo{However, we found the variance across random seeds to be negligible and not comparing across random seeds is a standard practice when working with large language models where the cost of a single run is significant.}
        \item Did you include the total amount of compute and the type of resources used (e.g., type of GPUs, internal cluster, or cloud provider)?
    \answerYes{In Appendix \ref{appendix:Hyperparameters} we report our hardware configuration.}
\end{enumerate}

\item If you are using existing assets (e.g., code, data, models) or curating/releasing new assets...
\begin{enumerate}
  \item If your work uses existing assets, did you cite the creators?
    \answerYes{In Appendix \ref{appendix:Hyperparameters}.}
  \item Did you mention the license of the assets?
    \answerYes{In Appendix \ref{appendix:Hyperparameters}.}
  \item Did you include any new assets either in the supplemental material or as a URL? \answerNA{}
  \item Did you discuss whether and how consent was obtained from people whose data you're using/curating?
    \answerNA{}
  \item Did you discuss whether the data you are using/curating contains personally identifiable information or offensive content?
    \answerNA{}
\end{enumerate}

\item If you used crowdsourcing or conducted research with human subjects...
\begin{enumerate}
  \item Did you include the full text of instructions given to participants and screenshots, if applicable?
    \answerNA{}
  \item Did you describe any potential participant risks, with links to Institutional Review Board (IRB) approvals, if applicable?
    \answerNA{}
  \item Did you include the estimated hourly wage paid to participants and the total amount spent on participant compensation?
    \answerNA{}
\end{enumerate}

\end{enumerate}

\appendix
\clearpage
\newpage
\newpage
\section{Broader impacts}
\label{section:Broader}
The focus area of this paper --- fine-tuning large language models --- is aligned with an important line of work on addressing the problem of social bias in large language models \citep{ShengCNP_LM_bias19,liang2021understanding}. As the training data for large language models consists mainly of crawled user-generated content, a number of factors (from crawling methodology to Internet participation inequalities and moderation practices) leads to an over-representation of certain viewpoints and voices exceeding their prevalence in the general population. This poses a risk of amplifying biases and harms through a language model perpetuating these voices \citep{Bender_parrots,blodgett-bias-survey,ShengCNP_LM_bias19,ethical_lm_deepmind,ethical_lm_deepmind2}. Numerous problems related to addressing data bias in language generation (e.g. controlling for gender distribution in generated texts) can be naturally posed as generative distributional control (GDC), the framework we focus our experiments on. The \emph{distributional} character of these data bias problems lies in the fact that desirable properties of generated texts are defined for a collection of samples, not only for individual samples. Our theoretical analyses of reward maximization and distribution matching approaches as well as our algorithmic improvements to the \GDC framework --- termed \GDCplus --- are therefore also a contribution to the problem of bias in language models. However, we need to be aware that \GDCplus, KL-control as well as controllable language generation techniques in general, can also be diverted to malicious uses such as spreading misinformation or generating harmful content.

\section{Extended Related Work}
\label{appendix:related}

\paragraph{Reinforcement learning for language generation} 
Most previous attempts at steering language models to conform to global constraints defined over entire sequences have employed reinforcement learning.
This includes using Reinforce~\citep{Williams92} for machine translation \cite{seq_lvl_train_RanzatoCAZ15}, actor critic~\citep{conda_actor} for abstractive summarization~\citep{PaulusXS18}, caption generation~\citep{RL_Img2txt_LiuZYG016}, dialogue~\citep{RL_dialogue_LiMRJGG16}, and video captioning~\citep{PasunuruB17}. Some approaches (for instance, in machine translation and summarization \citep{seq_lvl_train_RanzatoCAZ15, BahdanauBXGLPCB17}) directly optimize performance metrics such as BLEU and ROUGE at training time. 
Others use heuristic rewards (for instance \citet{RL_dialogue_LiMRJGG16} for dialogue generation and \citet{RL_TambwekarDMMHR19} for story generation) in order to obtain certain \textit{a priori} desirable features of generated sequences that then incentivize good performance on target metrics. 
 Catastrophic forgetting is a frequent problem of these fine-tuning approaches: reward maximization happens at the expense of large deviations from the original model. 
This problem is sometimes addressed by imposing a penalty term to the rewards, such as the KL divergence between the trained policy and the auto-regressive model. This approach, termed ``conservative fine-tuning", was applied to generating melodies with music theory rewards and organic molecules with synthesizability rewards by \citet{KL_Jaques17} as well fine-tuning language models for controllable language generation by \citet{Ziegler19}. This solution often has hard time balancing between the reward term and the KL penalty term, leading to instability in training ~\citep{khalifa_2021,pmlr-v162-korbak22a}. Unlike this approach, KL-DPG determines an optimal distribution that satisfies both requirements.

\paragraph{RM and DM objectives in control problems}

While RM is the dominant approach to tackling control problems \citep{Sutton2018} and is sometimes argued to be sufficient for any intelligent behavior \citep{SILVER2021103535}, prior work explored the benefits of alternative objectives formulated as DM: minimizing divergence from some target distribution $p$. Prominent examples of (families of) DM objectives 
include control state marginal matching \citep{smm2019} active inference \citep{friston2010action,BUCKLEY201755} and control-as-inference \citep{kappen2012optimal,todorov,levine2018reinforcement}. \cite{hafner2020action} propose a \emph{reverse} KL from a joint distribution over observations and latent variables as a universal objective for action and perception that --- depending on a choice of the target $p$ --- gives rise to many familiar objectives, including empowerment \citep{1554676}, maximum entropy RL \citep{pmlr-v70-haarnoja17a} or KL-control \citep{todorov}. In a similar vein, \cite{millidge2021understanding} compare RM and DM objectives (or, evidence and divergence objectives, according to their terminology) in the context of exploration. They conclude that information-seeking exploration arises naturally in DM but \emph{not} in RM. This is because, when the target distribution $p$ involves latent variables, a DM objective decomposes into an information gain term that pushes the agent to seek observations that are most informative of latent variables. In contrast, RM objectives entail \emph{minimizing} information gain between latent variables and observations. Finally, \citep{rl_kl_penalties} defend an interpretation of KL-control for controlling language models as Bayesian inference: updating a prior $a$ to conform to evidence provided by a reward function $R$.

\paragraph{Maximum entropy RL} Maximum entropy RL (MaxEnt RL)'s objective is maximising expected reward minus policy entropy. KL-control can be seen as generalisation of maximum-entropy RL \citep{pmlr-v70-haarnoja17a,sac} to informed priors. If in \eqref{RTZ_first_mention} we chose $a(x)$ to be a uniform distribution (an uninformed prior) instead of a pretrained LM distribution, then the KL penalty $\KL(\pit, a)$ would reduce to an entropy bonus and KL-control’s objective would reduce to a standard Maximum entropy RL objective. Both KL-control and Maximum entropy RL can be derived from a general framework of control-as-inference \citep{levine2018reinforcement} which poses control as minimising KL from a certain target distribution. However, while KL-control \citep{Ziegler19} and DPG directly minimise a single KL from a target distribution over whole sequences (trajectories), most practical algorithms in the maximum entropy family RL approximate it by related but importantly different objectives. 

The three biggest differences between MaxEnt RL on the one hand and DPG and KL-control \citep{Ziegler19} on the other hand are as follows:
\begin{enumerate}
    \item KL-control implicit target distribution $p_z$ and DPG’s target distribution $p$ are over whole sequences (trajectories) while in most MaxEnt RL algorithms the target distribution over actions conditioned on a state: $\pi^*(a|s)$. For instance in both SQL \citep{pmlr-v70-haarnoja17a} and SAC \citep{sac} the target distribution is defined as $\pi^*(a|s) = \exp(Q_\theta(s,a))/Z_\theta(s)$, where $Q$ is a state-action value function and $Z$ is a partition function of for a given state, both dependent on policy parameters $\theta$.
    \item  KL-control's implicit target distribution and DPG’s target distribution are predefined (i.e. held constant throughout training). In MaxEnt RL it typically undergoes updates. Again, in both SQL \citep{pmlr-v70-haarnoja17a} and SAC \citep{sac} they depend on a Q function which is continuously updated on new trajectories.
    \item KL-control's implicit target distribution $p_z$ and DPG's target distribution $p$ involve an informed prior $a(x)$: a pretrained language model. In most MaxEnt RL algorithms, the prior is assumed to be a uniform distribution.
\end{enumerate}

Because MaxEnt RL algorithms do not approximate a constant, predefined target distribution, they cannot be framed as minimising a single KL objective. Instead, they typically implement (soft) policy iteration \citep{Sutton2018}: they alternate between defining a new target distribution (policy evaluation) and minimising KL from that current target distribution (policy improvement). In other words, minimising KL is a subroutine of policy iteration, not an objective in itself.

Perhaps the closest method to KL-control and DPG in the larger family of inference-based RL \citep{coadaptation} is AWR \citep{awr}, which minimises the \emph{forward} KL from a target distribution $\frac{1}{Z}\mu(a|s)\exp(A(s,a))$, where $\mu$ is a behavioural policy implicitly defined by the trajectory buffer and $A$ is the advantage.  Here, the prior is informative and given by the policy from a previous iteration $k$. However, the target distribution is not constant: it is updated on each iteration.

\paragraph{State marginal matching} State marginal matching \citep{maxentexplor,smm2019} is an approach to exploration in RL. It poses exploration as learning a policy $\pi$ that induces a state marginal distribution $\rho_\pi(s) = \E \sum_{t=1}^T \mathds{1} (s_t=t)$ that matches a given target state distribution $p^*$. While this approach differs in motivation from DPG and KL-control (it solves the problem of exploration in the space of policies, not constraint satisfaction), it optimises a similar divergence objective: $\KL(\pi, p^*)$. Unlike in maximum-entropy RL, the target $p^*$ is fixed. However, $p^*$ is a distribution over states, not trajectories (as in the case of $p$ in DPG and $p_z$ in KL-control). There is no obvious notion of state in the controllable language generation tasks we consider other than treating the whole sequence as a state. 

\paragraph{Baselines in Reinforcement Learning}

In the context of reinforcement learning, baselines were introduced by \citet{suttonphd}. \citet{williams1987,Williams92} has shown them to reduce variance in a number of use cases and also proved that they do not introduce bias. \citet{dayan_baseline} was the first to observe and confirm experimentally that the optimal constant baseline is not equal to expected reward in a simple two-arm bandit setting. This result was generalized to POMDPs (Partially Observable Markov Decision Processes) by \citet[section 3.1.3, p. 540]{weavertao} and variable baselines by \citet[theorem 13, p. 1489]{Greensmith} who also proved bounds on the variance of gradient estimates. The optimal baseline, however, is rarely used in practice (\citet{Sutton2018}; for an exception, see \citep{PETERS2008682}). Outside RL, baselines were also used in the context of learning inference networks for amortized variational inference by \citet{mnih_nvil} and found to yield similar variance reduction.

\paragraph{Energy-based models for language}
Energy-based models (EBMs)~\citep{Hinton02,lecun_tutorial_2006,RanzatoBCL07} are a family of models in which learning and inference are done by associating an unnormalized probability with each configuration of observed and latent variables. Early examples of EBMs applied to natural language processing include sequence labeling problems (e.g. tagging) exploiting global properties of a  sequence~\citep{andor_globally_2016,Belanger:2016:SPE:3045390.3045495}. The recent surge of interest in EBMs has not left natural language processing unaffected (see \cite{Bakhtin2020EnergyBasedMF} for a survey). \citet{Tu2020ENGINEEI} proposed an energy-based inference networks for non-autoregressive machine translation while \citet{Naskar2020EnergyBasedRI} use an EBM for reranking candidate translations according to their predicted BLEU scores. \citet{A-parshakova-etal-2019-global} and \citet{Deng_EBM_20} augment an autoregressive language models with an additional global factor to obtain a lower perplexity on the training data. \citet{ClarkLLM20} poses non-autoregressive language modeling as training an energy-based cloze task scorer using noise-contrastive estimation \citep{nce}. \citet{he-etal-2021-joint} obtain better calibration on natural language inference tasks by augmenting and training the classifier jointly with an energy-based model modeling the marginal distribution over samples, again using noise-contrastive estimation. In consequence, the classifier tends to assign more conservative (high-entropy) predictions to high-energy (less likely, possibly out of distribution) samples. 

\section{Additional proofs}
\label{appendix:baselines}

\subsection{Optimal baselines in RL}
\label{appendix:optimal-baseline}
\fmd{(4-10) I have corrected some typos in the text and tried to simplify a bit one long derivation.}
Despite its widespread use, the baseline as mean of reward 
\begin{equation}
    \label{eq:common-baseline}
    B^{\text{RL}} = \E_{x \sim \pi_\theta(x)} R(x)
\end{equation}
is not the optimal constant baseline for reward maximization objectives in RL. The optimal constant baseline, i.e. one yielding the minimal variance of the gradient, is given by:
\begin{equation}
    \label{eq:opt-baseline}
    B^{*} = \frac{\E_{x \sim \pit}[R(x)\left(\nabt \log \pit(x) \right)^2]}{\E_{x \sim \pit} [\left(\nabt \log \pit(x) \right)^2]}.
\end{equation}

In order to maintain accessibility, in this section, we provide a self-contained derivation of this optimal form of baselines \eqref{eq:opt-baseline} and and connect it to the commonly used form \eqref{eq:common-baseline}.\footnote{The formula for the optimal baseline in \eqref{eq:opt-baseline} was originally proved by \citet{weavertao} but here we provide a simpler proof sketched by Sergey Levine in his slides:  \url{http://rail.eecs.berkeley.edu/deeprlcourse-fa17/f17docs/lecture_4_policy_gradient.pdf}}

First, recall that $R(x)$ is a reward associated with an input $x$. $B$ is a baseline value subtracted from the reward that does not introduce bias in gradient estimation. Now let's denote the gradient wrt an individual sample $x$ as $\grad(x)$ where
\begin{equation}
    \grad(x) = [ R(x) - B ] \nabt \log \pit(x),
\end{equation}
and the estimate of the gradient as
\begin{equation}
    \gradest = \E_{x \sim \pit} \grad(x).
\end{equation}

Using the general identity $\mathbf{var}(z) = \E [z^2] - [\E z]^2$, the variance of the gradient takes the form:
\begin{equation}
    \vargrad = \E_{x \sim \pit} [\grad(x)^2] - \gradest^2
\end{equation}

Now let's take the gradient of this variance with respect to $B$ and solve to find the baseline form with minimal variance:
\begin{align}
   \label{eq:grad_var_grad}
    \frac{d\vargrad}{dB} &= 
    \frac{d}{dB} \E_{x \sim \pit} [(\grad(x))^{2}] - \frac{d}{dB} (\E_{x \sim \pit} [\grad(x)])^2.
\end{align}
The second term of the \ptmd{right}
hand side of \eqref{eq:grad_var_grad} is equal to zero, since $B$ does not introduce bias into $\gradest$: 
\begin{align*}
    \frac{d}{dB} \left(\E_{x \sim \pit} [\grad(x)]\right)^2 
    &= \frac{d}{dB} \left(\E_{x \sim \pit} \left[ (R(x) -B) \nabt \log \pit(x)  \right] \right)^2 \\
    &= \frac{d}{dB} \left(\E_{x \sim \pit} \left[ R(x) \nabla \log \pit(x) \right]\right)^2 = 0.
\end{align*}

Plugging this back into \eqref{eq:grad_var_grad}, we obtain:
\begin{align*}
    \frac{d\vargrad}{dB}  
  &= \frac{d}{dB} \E_{x \sim \pit} [(\grad(x))^{2}] \\
&= \E_{x \sim \pit} \left[ \frac{d}{dB} \left[\left(R(x)^2 + B^2 - 2R(x)B\right) \left(\nablapitlog \right)^2  \right]\right] \\
&= \E_{x \sim \pit} (2B- 2R(x))  (\nablapitlog)^2 \\
&= 2B\ \E_{x \sim \pit} (\nablapitlog)^2 - 2\ \E_{x \sim \pit} R(x) \left(\nablapitlog \right)^2.
\end{align*}

Then, solving $\frac{d\vargrad}{dB} = 0$ for $B$, we obtain the optimal form of the baseline $B^{*}$ as required:
\begin{equation}
    B^{*} = \frac{\E_{x \sim \pit} [R(x)\left(\nablapitlog \right)^2]}{\E_{x \sim \pit} [\left(\nablapitlog \right)^2]}. 
\end{equation}

This can be interpreted as average reward (as in $B^{\text{RL}}$) but weighted by gradient magnitudes $(\nabt \log \pit(x))^2$. 
Moreover, $B^{*} = B^{\text{RL}}$ is recovered \textbf{under the condition that} the reward $R(x)$ is uncorrelated (\textit{a fortiori} independent) from $(\nabt \log \pit(x))^2$. If that were the case, we would have:
\begin{align}
    B^{*} &= \frac{\E_{x \sim \pit}[R(x)\left(\nabt \log \pit(x) \right)^2]}{\E_{x \sim \pit} [\left(\nabt \log \pit(x) \right)^2]} \\
    &= \frac{\E_{x \sim \pit} [R(x)] \; \E_{x \sim \pit} [\left(\nablapitlog\right)^2]}{\E_{x \sim \pit} [\left(\nablapitlog \right)^2]}  \\
    &= \E_{x \sim \pit} [R(x)] = B^{\text{RL}}.
\end{align}
%
\subsection{unbiasedness of PG baseline}
Baselines are a standard variance reduction technique in the context of Policy Gradients \citep{Sutton2018}. The idea is to subtract from the reward $R(x)$ a value $B$ 
that does not introduce bias to the gradients but may change variance. Equation \eqref{eq:REINFORCE} then takes the following form:
\begin{equation}
\nabt \EX{\pit} R(x) = \EX{\pit} (R(x)-B)\, \nabt \log \pit(x).
\end{equation}

To see that $B$ does not introduce bias, we can rewrite \eqref{pg_baseline} as:
\begin{equation}
\label{pg_baseline_split}
\EX{x \sim \pit} R(x) \nabt \log \pit(x) - B\, \EX{\pit} \nabt \log \pit(x)
\end{equation}
and note that the second term is null because $\sum_x \pit(x) \nabt \log \pit(x) = \nabt \sum_x \pit(x) = 0$. 

\subsection{Unbiasedness of \DPG Baseline}

Recall that the gradient estimate for DPG \citep{A-parshakova-etal-2019-global} has the following form:
\begin{equation}
    \E_{x \sim \pit} \frac{P(x)}{\pit(x)} \nabla_{\theta} \log \pit(x)
\end{equation}
After subtracting a baseline $B = Z$, it becomes
\begin{align}
    \E_{x \sim \pit} \Big[ \frac{P(x)}{\pit(x)} - Z \Big] \nabla_{\theta} \log \pit(x)
    &= \E_{x \sim \pit} \frac{P(x)}{\pit(x)} \nabla_{\theta} \log \pit(x) 
    -Z \Big[ \E_{x \sim \pit}  \nabla_{\theta} \log \pit(x) \Big]\\
    &= \E_{x \sim \pit} \frac{P(x)}{\pit(x)} \nabla_{\theta} \log \pit(x)
    - Z \Big[\sum_{x} \nabla_{\theta} \pit(x) \Big]
\end{align}
Here, the second term does not introduce bias because $Z \Big[\sum_x \nabt \pit(x) \Big]= 0$, leaving us with the same exact form of gradient as in the DPG algorithm.

\subsection{Unbiasedness of \DPGoff baseline}

Offline DPG, the off policy variant of DPG proposed in ~\cite{opt-rl-arxiv-2019,khalifa_2021} has the following gradient estimate:
\begin{equation}
    \E_{x \sim q}  \frac{P(x)}{q(x)} \nabla_{\theta} \log \pit(x) 
\end{equation}
Where $q$ is a proposal distribution (another auto-regressive model) used to detach the training of $\pit$ from the sampling process and allow more stable training.

Recall that the Baseline of \DPGoff is of the form:
\begin{equation}
    \Boff = Z\frac{\pit(x)}{q(x)},
\end{equation}
The $\frac{\pit(x)}{q(x)}$ term is an importance weight correcting for the bias introduced by sampling from~$q$.

\paragraph{Unbiasedness} To show that subtracting a baseline $\Boff = Z\frac{\pit(x)}{q(x)}$ doesn't introduce bias, let's rewrite the gradient estimate with added baseline as a sum of two terms:
\begin{align}
 \E_{x \sim q} \Big[ \frac{P(x)}{q(x)} - Z\frac{\pit(x)}{q(x)}\Big] \nabla_{\theta} \log \pit(x) 
    &= \Big[ \E_{x \sim q} \frac{P(x)}{q(x)} \nabla_{\theta} \log \pit \Big]
    - \Big[ \E_{x \sim q} Z\frac{\pit(x)}{q(x)} \nabla_{\theta} \log \pit \Big] \\
    &= \Big[ \E_{x \sim q} \frac{P(x)}{q(x)} \nabla_{\theta} \log \pit \Big]
    - Z \Big[ \sum_x \nabla_\theta \pit(x) \Big] 
\end{align}
Here again the second term does not introduce bias because  $Z \Big[ \sum_x \nabt \pit (x)\Big] = 0$. 

\paragraph{Null Advantage on Average} In the case of sampling with $\pit$ in the online DPG 
choosing $B=Z$ had the benefit that the advantage $R_\theta(x) - B$ was centered around $0$, namely: $\E_{x\sim \pit} [R_\theta(x) - Z] = 0$.

With the $\Boff$ baseline for the \DPGoff this important property is also maintained. The advantage now takes the form $ \frac{P(x)}{q(x)} - Z\frac{\pit(x)}{q(x)}$ and then:
\begin{align}
    \E_{x\sim q} \Big[\frac{P(x)}{q(x)} - Z\frac{\pit(x)}{q(x)}\Big] 
    &= \sum_x P(x) - Z \pit(x)\\
    &= Z - Z \sum_x \pit(x) = 0.\label{eq:null_advantage_in_avg}
\end{align}

\bigskip

To visualize things better, we elaborate the difference in forms of rewards, baseline and gradients before and after addition of the baseline between DPG (on policy) and \DPGoff (off policy) in Table \ref{tab:my_label}.



\begin{table*}[h]
    \centering
\begin{tabular}{l c c }
\toprule
 & \textbf{DPG} & \textbf{\DPGoff} \\
\midrule
 \textbf{Reward} & 
 $\frac{P(x)}{\pi_\theta(x)}$ & 
 $\frac{P(x)}{q(x)}$\\
 \addlinespace
 \textbf{$\nabt$} & 
{\small $\E_{x \sim \pit} \frac{P(x)}{\pi_\theta(x)} \nabt \log \pit(x)$} & 
{\small $\E_{x \sim q} \frac{P(x)}{q(x)} \nabt  \log \pit(x)$} \\ 
 \addlinespace
 \textbf{Baseline}& 
 $Z$   & 
 $Z\frac{\pit(x)}{q(x)}$\\
 \addlinespace
 \addlinespace
 \textbf{Advantage} & 
 $\frac{P(x)}{\pi_\theta(x)} - Z$ & 
 $\frac{P(x)}{q(x)} - Z\frac{\pit(x)}{q(x)}$ \\
 \addlinespace
 \textbf{$\nabt$ with baseline} & 
 {\small $\E_{x \sim \pit} \Big[ \frac{P(x)}{\pi_\theta(x)} - Z \Big] \nabt \log \pit(x)$} & 
{\small $\E_{x \sim q} \Big[ \frac{P(x)}{q(x)} - Z\frac{\pit(x)}{q(x)} \Big] \nabt  \log \pit(x)$} \\
 \bottomrule
\end{tabular}
    \caption{\small{A comparison of Online DPG and Offline DPG (\DPGoff)} forms of Reward, Baseline, Advantage, and Gradient of the loss function (the PG-term) before ($\nabla_\theta$) and after ($\nabla_\theta$ with Baseline) including a baseline for variance reduction.}
    \label{tab:my_label}
\end{table*}

\newpage
\section{Additionals details on metrics and Algorithms}
\label{detailed-metrics}

Calculation of metrics relative to $p$, such as $\KL(p,\pi_\theta)$, is not straightforward since the distribution $p \propto P$ is only implicitly represented by the unnormalized EBM $P$, and one cannot easily obtain direct samples from $p$. Instead, we apply the following workarounds. Given $P$ and a proposal distribution $q$ that we can sample from, using importance sampling \citep{owen_chapter_importance_sampling_2013}, we calculate the partition function $Z$ as follows: 
\begin{align} 
            Z &= \sum_x P(x) = \sum_x q(x)\ P(x)/q(x)\\
                    &= \mathbb{E}_{x\sim q}\ P(x)/q(x).
\end{align}

The precision of this estimate depends on the sample size and the quality of the proposal distribution $q$. We calculate a moving average estimate $Z_\text{MA}$ of $Z$ which is then used inside the estimations of $\KL(p, \pit)$ and $\KL(p, q)$ (see below Algorithm~\ref{appendix:al:DPG}, lines 7 and 8). 
$Z_\text{MA}$ is updated at each training iteration. $Z_\text{MA}$ is an unbiased estimate of $Z$ because each $\hat{Z}_i$ is an unbiased estimate of $Z$ based on $K$ samples. Moreover, because the proposal distribution $q$ evolves and gets closer to the target distribution $p$, the quality of the  estimate of $Z_\text{MA}$ through importance sampling increases.

With an estimate of $Z$, we can compute $\KL(p, \pit)$ as
\begin{align}
\KL(p, \pit) &= \sum_x p(x) \log \frac{p(x)}{\pit(x)} \\
&= \sum_x p(x) \log \frac{P(x)}{Z \pit(x)} \\
&= -\log Z + \sum_x p(x) \log \frac{P(x)}{\pit(x)} \\
&= -\log Z + \sum_x q(x) \frac{p(x)}{q(x)} \log \frac{P(x)}{\pit(x)} \\
&= -\log Z + \frac{1}{Z} \mathbb{E}_{x\sim q} \frac{P(x)}{q(x)} \log \frac{P(x)}{\pit(x)}.
\end{align}
Similarly, for $\TVD(p,  \pit)$:
\begin{align}
            \TVD(p, \pit) &= \frac{1}{2} \sum_x |p(x)-\pit(x)| \\
                            &= \frac{1}{2} \sum_x q(x)\ \left|\frac{\pit(x)}{q(x)} - \frac{p(x)}{q(x)}\right| \\
                            &= \frac{1}{2} \sum_x q(x)\ \left|\frac{\pit(x)}{q(x)} - \frac{P(x)}{Z\ q(x)}\right|\\
                            &= \frac{1}{2} \mathbb{E}_{x\sim q}\ \left|\frac{\pit(x)}{q(x)} - \frac{P(x)}{Z\ q(x)}\right|.
\end{align}

See Algorithm \ref{appendix:al:DPG} for a detailed pseudocode describing how metric computation is integrated in the training loop of KL-DPG.
\begin{algorithm*}[h]
\caption{\ KL-DPG with baseline (detailed) \label{appendix:al:DPG}}
\begin{algorithmic}[1]
\Require $P$, initial policy $q$
\State $\pi_\theta \gets q$
\State $Z_\text{MA} \gets 0$                               
\For{each iteration $i$}
\For{each step  $k\in[1,K]$}
    \State sample $x_k$ from $q(\cdot)$
    \State $\theta \gets \theta + \alpha^{(\theta)} \Big[ \frac{P(x_k)}{q(x_k)} - Z\frac{\pit(x_k)}{q(x_k)} \Big] \nabla_\theta \log \pi_\theta(x_k)$ 
\EndFor
\State $\hat{Z}_i \leftarrow \frac{1}{K} \sum_k P(x_k)/q(x_k)$
\State $Z_\text{MA} \gets \frac{i * Z_\text{MA}+\hat{Z}_i}{i + 1}$ 
\State $\KLhat(p, \pit) \leftarrow  
 -\log Z_\text{MA} + 1/(KZ_\text{MA}) \sum_k  \frac{P(x_k)}{q(x_k)} \log \frac{P(x_k)}{\pit(x_k)}$

\State $\KLhat(p,q) \leftarrow  
 -\log Z_\text{MA} + 1/(KZ_\text{MA}) \sum_k  \frac{P(x_k)}{q(x_k)} \log \frac{P(x_k)}{q(x_k)}$

\If{$\KLhat(p, \pit) <  \KLhat(p,q)$}
    \State $q \gets \pi_\theta$
\EndIf
\EndFor
\Ensure $\pi_\theta$
\end{algorithmic}
\end{algorithm*}

\newpage
\section{Hyperparameters and training details}
\label{appendix:Hyperparameters}

We implemented all models using PyTorch \citep{pytorch} and HuggingFace ~\citep{huggingface}. Based on \cite{khalifa_2021} source code published under CC BY-NC-SA 4.0 license: \url{https://github.com/naver/gdc}. The two pretrained models used in our experiments are available on Hugginface Model Hub: \texttt{gpt}\footnote{\url{https://huggingface.co/gpt2}} and \texttt{mkhalifa/gpt2-biographies}.\footnote{\url{https://huggingface.co/mkhalifa/gpt2-biographies}} Each training run took approximately 5 days on 2 Nvidia V100 GPUs. For a detailed list of hyperparameter values, see Table \ref{table:hyperparams}; for a description of hyperparameters specific to Ziegler and GDC, see \citep{Ziegler19} and \citep{khalifa_2021}. 

\begin{table}[H]
    \footnotesize
    \centering
    \begin{tabular}{ll}
    \toprule
    \textbf{Hyperparameter} & \textbf{Value}  \\
    \toprule
    \multicolumn{2}{c}{\textbf{Common}} \\
    batch size & 512 \\
    sequence length & 40 tokens \\
    learning rate & $1.41 \times 10^{-5}$ \\
    dropout rate & 0.1 \\
    optimizer & Adam \citep{kingma2014adam}\\
    warmup epochs & 100 \\
    total epochs & 4500 \\
    base LM & GPT-2 small (117M params) \\
    \multicolumn{2}{c}{\textbf{GDC}} \\
    sample size for learning $\lambda$ & 10240 \\
    learning rate for $\lambda$ & 0.5 \\
    tolerance for $\lambda$ & 0.01 \\
    \multicolumn{2}{c}{\textbf{Ziegler}} \\
    $\gamma$ & 1 \\
    $\lambda$ & 0.95 \\
    clip range & 0.2 \\
    target KL & 6.0 \\
    initial KL coefficient & 0.2 \\
    horizon & $10^4$ \\
    \bottomrule
    \vspace{5px}
    \end{tabular}
    \caption{Hyperparameters used throughout all experiments.}
    \label{table:hyperparams}
\end{table}
\newpage 
\section{Extended evaluation (Table View)}

\begin{table}[H]
\tiny
\centering
\begin{tabular}{llcccccccc}
\toprule
 &   &  &  \multicolumn{2}{c}{\textbf{Fluency}} & \multicolumn{3}{c}{\textbf{Sentence Level Diversity}} &  \multicolumn{2}{l}{\textbf{Corpus Level Diversity}} \\
& \textbf{Method}  &  \textbf{Ctrl ($\uparrow$)} & \textbf{KL{(p,$\boldsymbol{\pi}$)}} ($\downarrow$) &  \textbf{KL(pi,a)} ($\downarrow$)&  \textbf{Dist-1} ($\uparrow$)&  \textbf{Dist-2} ($\uparrow$)&  \textbf{Dist-3} ($\uparrow$)&  \textbf{SB-4} ($\downarrow$)&  \textbf{SB-5}($\downarrow$) \\
\midrule
\multicolumn{10}{c}{\textbf{Pointwise Constraints Experiments}}\\
\midrule
Word Amazing 
 & \gc{Original LM} &  \gc{0.00} &  \gc{6.02} &  \gc{0.00} &  \gc{0.86} &  \gc{0.94} &  \gc{0.92} &  \gc{0.89} &  \gc{0.82} \\
 & Reinforce &  \r{1.00} &  \r{134.31} & \r{78.39} &  \r{0.69} &  \r{0.91} &  \r{0.94} &  \r{0.98} &  \r{0.96} \\
 & Ziegler &  \textbf{0.82} &  4.56 &  5.88 &  0.86 &  0.95 &  0.94 &  0.94&  0.88 \\
 & GDC &  0.65 &  2.57 &  5.06 &  0.86 &  0.95 &  0.94 &  0.93 &  0.87 \\
 & GDC++ (Ours)&  \underline{0.69} &  \underline{\textbf{2.10}} &  \underline{\textbf{4.74}} &  \underline{\textbf{0.87}} &  0.95 &  0.94 &  0.93 &  0.87 \\
   \midrule
Word WikiLeaks 
 & \gc{Original LM} &  \gc{0.00} &  \gc{8.54} &  \gc{0.00} &  \gc{0.86} &  \gc{0.9}4 &  \gc{0.92} &  \gc{0.89} &  \gc{0.80} \\
 & Reinforce &  \r{1.00}&  \r{8.00} &  \r{117.24} &  \r{0.38} &  \r{0.56} & \r{0.64} & \r{0.98} &  \r{0.97} \\
 & Ziegler &  0.68 &  0.00 &  6.03 &  0.87 &  0.96 &  0.94 &  0.95 &  0.90 \\
 & GDC &  0.75 &  3.22 &  7.96 &  0.88 &  0.96 &  0.94 &   0.95 &  0.90 \\
 & GDC++ (Ours) &    \underline{\textbf{0.77}} &  \underline{\textbf{2.21}} &  \underline{\textbf{7.53}} &  0.88 &  0.96 &  0.94 &  0.95 &  0.91 \\
  \midrule
Wordlist Science 
 & \gc{Original LM} &  \gc{0.06} & \gc{2.79} & \gc{0.00} &  \gc{0.86} &  \gc{0.94} &  \gc{0.92} &  \gc{0.89} &  \gc{0.81} \\
 & Reinforce&  \r{1.00} &  \r{140.02} &  \r{66.68} &  \r{0.29} &  \r{0.41} &  \r{0.49} & \r{0.98} &   \r{0.97} \\
  & Ziegler &  \textbf{1.00} &    6.1 &  5.88 &  0.86 &  0.95 &  0.93 &  0.95 &  0.90\\
 & GDC &  0.52 &  2.27 &  2.89 &  0.86 &  0.95 &  0.93 &  0.93 &  0.87 \\
 & GDC++ (Ours) &  \underline{0.54} &  \underline{1.78} &  \underline{\textbf{2.11}} &  0.86 &  0.95 &  0.93 & \underline{\textbf{0.92}} &  \underline{\textbf{0.86}} \\
  \midrule
Wordlist Politics 
 & \gc{Original LM} &  \gc{0.07} &  \gc{2.65} &  \gc{0.01} &  \gc{0.86} &  \gc{0.94} &  \gc{0.92} &  \gc{0.89} &  \gc{0.81} \\
 & Reinforce&  \r{1.00} &  \r{263.79} &  \r{65.06} &  \r{0.26} &  \r{0.40} &  \r{0.51}& \r{0.98} &  \r{0.97} \\
 & Ziegler &  \textbf{1.00} &  8.46 &  5.92 &  0.87 &  0.96 &  0.94 &  0.96 &  0.92 \\
 & GDC &  \underline{0.58} &  2.70 &  2.49 &  0.87 &  \underline{0.96} &  \underline{0.94} &  0.93 &  0.88 \\
 & GDC++ (Ours) &  0.49 &  \underline{\textbf{2.01}} &  \underline{\textbf{1.35}} &  0.87 &  0.95 &  0.93 &  0.93 &  \underline{\textbf{0.87}} \\
  \midrule
+ve Sentiment 
 & \gc{Original LM} &  \gc{0.17} &  \gc{2.06} &  \gc{0.01} &  \gc{0.86} &  \gc{0.94} &  \gc{0.93} &  \gc{0.89} &  \gc{0.81} \\
 & Reinforce &  \r{1.00} &  \r{153.75} &  \r{80.07} &  \r{0.27} &  \r{0.37} &  \r{0.41} &  \r{0.97} &  \r{0.95} \\
 & Ziegler &  \textbf{0.98} &  5.70 &  5.98 &  0.85 &  0.96 &  0.94 &   0.96 &  0.91 \\
 & GDC &  0.59 &  1.68 &  1.89 &  0.86 &  0.95 &  0.94 &  0.93 &  0.87 \\
 & GDC++ (Ours) &  \underline{0.60} &  \underline{\textbf{1.67}} &  \underline{\textbf{1.88}} &  0.86 &  0.95 &  0.94 &  0.93 &  0.87 \\
  \midrule
-ve Sentiment 
 & \gc{Original LM} &  \gc{0.13} &  \gc{2.14} &  \gc{0.01} &  \gc{0.86} &  \gc{0.94} &  \gc{0.92} &   \gc{0.90} &  \gc{0.82} \\
 & Reinforce &  \r{1.00} &  \r{88.48} &  \r{70.38} &  \r{0.83} &  \r{0.96} &  \r{0.94} &  \r{0.97} &  \r{0.93} \\
 & Ziegler &  0.95 &  6.12 &  6.00 &  0.84 &  0.95 &  0.94 &  0.96 &  0.92 \\
 & GDC &  \underline{0.52} &  1.72 &  1.79 &  0.86 &  0.95 &  0.94 &  0.94 &  0.88 \\
 & GDC++ (Ours) &  0.51 &  \underline{\textbf{1.66}} &  \underline{\textbf{1.63}} &  0.86 &  0.95 &  0.94 & \underline{\textbf{0.93}} &  0.88 \\
  \midrule
  \multicolumn{10}{c}{\textbf{Distributional Constraints Experiments}}\\
  \midrule
Single 
 & \gc{Original LM} &  \gc{0.19} &  \gc{0.39} &  \gc{0.01} &  \gc{0.90} &  \gc{0.95} &  \gc{0.92} &  \gc{0.94} &  \gc{0.90} \\
 & GDC &  0.80 &  0.74 &  0.71 &  0.89 &  0.95 &  0.92 &  0.95 &  0.90 \\
 & GDC++ (Ours) &  \underline{\textbf{0.81}} &  \underline{\textbf{0.33}} &  \underline{\textbf{0.66}} &  0.89 &  0.95 &  0.92 &  \underline{\textbf{0.94}} &  0.90 \\
  \midrule
Multiple 
 & \gc{Original LM} &  \gc{0.49} &  \gc{0.40} &  \gc{0.00} &  \gc{0.90} &  \gc{0.95} &  \gc{0.92} &  \gc{0.94} &  \gc{0.90} \\
 & GDC &  0.92 &  0.53 &  0.85 &  0.90 &  0.95 &  0.92 &  0.95 &  0.90 \\
 & GDC++ (Ours) &  \underline{\textbf{0.95}} &  \underline{\textbf{0.30}} &  \underline{\textbf{0.76}} &  0.90 &  0.95 &  0.92 &  0.95 &  0.90 \\
  \midrule
Hybrid Sports 
 & \gc{Original LM} &  \gc{0.22} &  \gc{0.20} &  \gc{0.00} &  \gc{0.90} &  \gc{0.95} &  \gc{0.92} &  \gc{0.94} &  \gc{0.90} \\
 & GDC &  \underline{\textbf{0.87}} &  \underline{\textbf{0.24}} &  2.65 &  0.93 &  0.95 &  0.92 &  0.96 &  0.92 \\
 & GDC++ (Ours) &  0.85 &  0.87 &  \underline{\textbf{2.35}} &  0.93 &  0.95 &  0.92 &  0.96 &  0.92 \\
  \midrule
Hybrid Science 
 & \gc{Original LM} &  \gc{0.09} &  \gc{0.00} &  \gc{0.00} &  \gc{0.90} &  \gc{0.95} &  \gc{0.92} &  \gc{0.94} &  \gc{0.89} \\
 & GDC &  0.68 &  1.52 &  3.92 &  0.88 &  0.95 &  0.91 &  0.95 &  0.92 \\
 & GDC++ (Ours) &  \underline{\textbf{0.70}} &  \underline{\textbf{1.41}} &  \underline{\textbf{3.83}} &  0.88 &  0.95 &  \underline{\textbf{0.92}} &  0.95 &  \underline{\textbf{0.91}} \\
\bottomrule
\vspace{5px}
\end{tabular}
\caption{
\small{
      Evaluation over 6 pointwise constraints experiments (tasks 1-6) and 4 distributional constraints experiments (tasks 7-10) for policies obtained from GDC\texttt{++} (ours), GDC, Ziegler and Reinforce.
      See figures \ref{fig:pointwise-compare-methods-split1}-\ref{fig:distributional-compare-methods-split} in the Appendix for a detailed view on each experiment.
      Results of the initial policy (\gc{Original LM}) are displayed for reference.The best method (excluding ties) overall is highlighted in \textbf{bold}, while the best method between GDC and GDC++ is \underline{underlined}. Runs that suffer degeneration due to \r{catastrophic forgetting} (measured by sequence level repetitions) are highlighted in red and excluded from best method comparison. Our method GDC++ that includes a baseline for variance reduction, outperforms GDC \citep{khalifa_2021} in 7/10 tasks in terms of control satisfaction rate (\textbf{Ctrl}), as well as convergence towards the optimal policy (\textbf{KL{(p,$\boldsymbol{\pi}$)}}) and distance from the original LM (\textbf{KL(pi,a)}) in 10/10 of the tasks. 
   }
   }
   \label{tab:all_experiments_results}
\end{table}


\begin{figure*}[h]
    \centering
    \includegraphics[width=\linewidth]{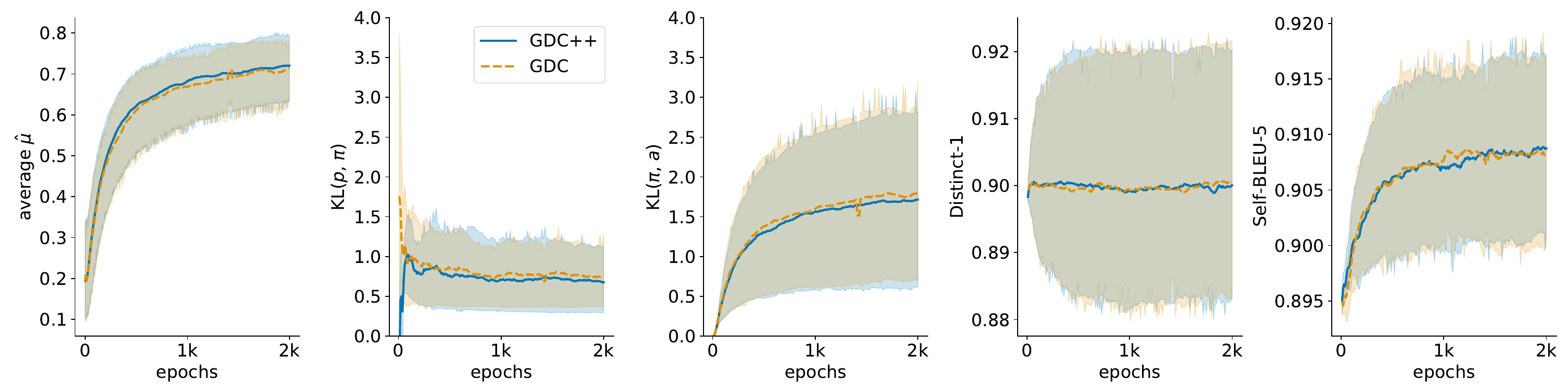} 
    \caption{\small{Evaluation metrics: average $\hat{\mu}$ ($\uparrow$ better), $\KL(p|\pi_{\theta})$ ($\downarrow$ better), $\KL(\pi_{\theta}|a)$ ($\downarrow$ better), Self-BLEU-5 ($\downarrow$ better), and Distinct-1 ($\uparrow$ better) on \textbf{aggregated} four  distributional constraints experiments:  
    \textbf{Task 7:} a single distributional constraint, \textbf{Task 8} and \textbf{Task 9:} a two hybrid constraint pairs, \textbf{Task 10:} Multiple Distributional constraints. For policies obtained from GDC\texttt{++} and GDC. Average $\hat{\mu}$ was computed for each experiment by mapping $\E_{x \sim q} \phi_i(x)$ for each constraint $i$ onto a $[0, 1]$ interval and averaging over constraints. See Figures \ref{fig:distributional-compare-methods-mu}-\ref{fig:distributional-compare-methods-split} in for a detailed view on each experiment.}}
    \label{fig:distributional-compare-methods-metrics}
\end{figure*}

\begin{figure*}[h]
    \centering
    \begin{tabularx}{\textwidth}{p{0.15\textwidth} p{0.28\textwidth} p{0.28\textwidth} p{0.2\textwidth}}
  & (a) & (b) & (c) \\
  \end{tabularx}
    \includegraphics[width=\linewidth]{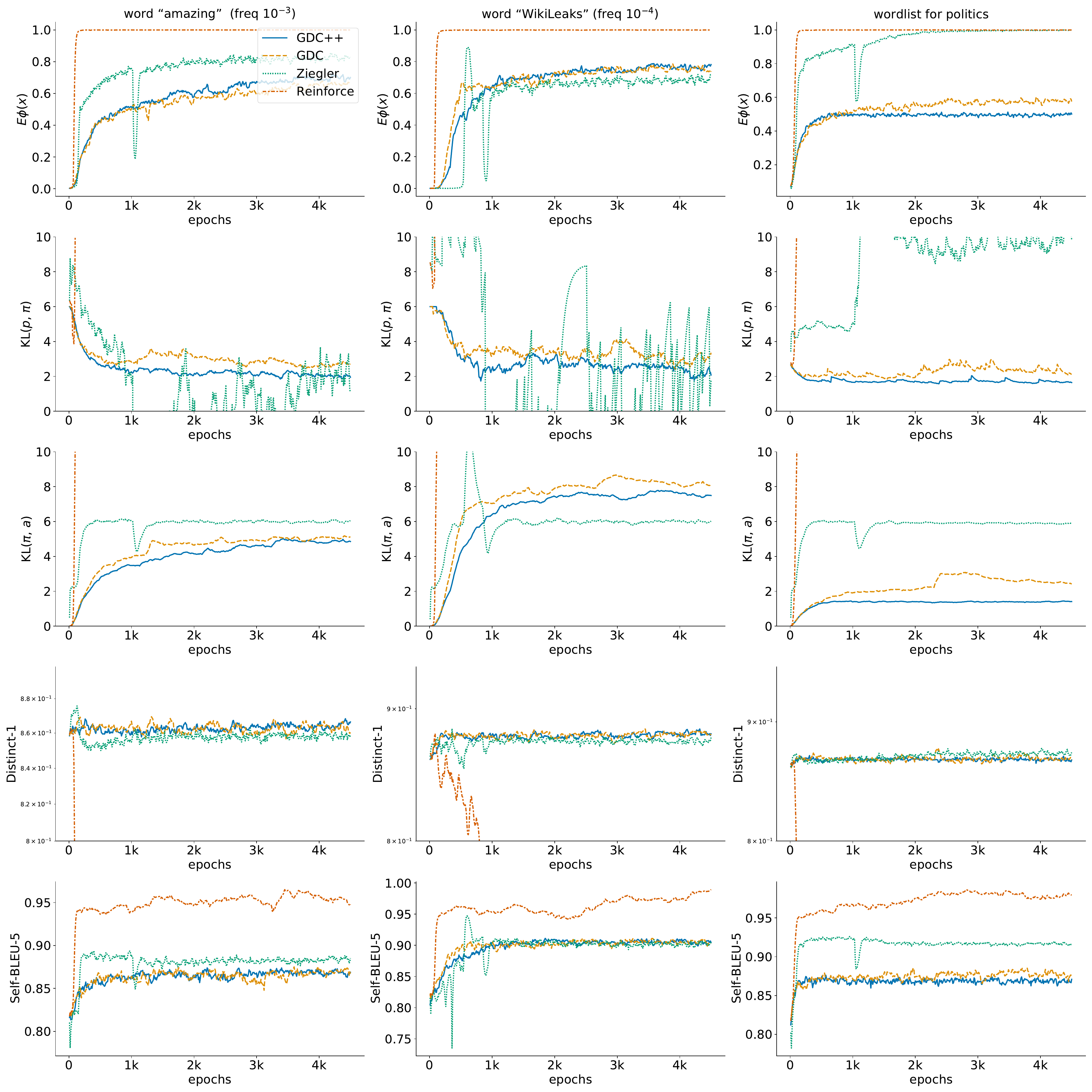} 
    \caption{\small{Evaluation metrics $\mathbb{E}_{\pit} \phi(x)$, $\text{KL}(p|\pi_{\theta})$ ($\downarrow$ better), $\text{KL}(\pi_{\theta}|a)$ ($\downarrow$ better), Self-BLEU-5 ($\downarrow$ better), and Distinct-1 ($\uparrow$ better) for three constraints types: \textbf{Task 1: Word "amazing"} Fig.(a), \textbf{Task 2: Word "wikileaks"} Fig.(b) and \textbf{Task 3: Wordlist "politics"} Fig.(c) for policies obtained from GDC\texttt{++}, GDC, Ziegler and Reinforce.}}
    \label{fig:pointwise-compare-methods-split1}
\end{figure*}

\begin{figure*}[h]
    \centering
        \begin{tabularx}{\textwidth}{p{0.15\textwidth} p{0.28\textwidth} p{0.28\textwidth} p{0.2\textwidth}}
  & (a) & (b) & (c) \\
  \end{tabularx}
    \includegraphics[width=\linewidth]{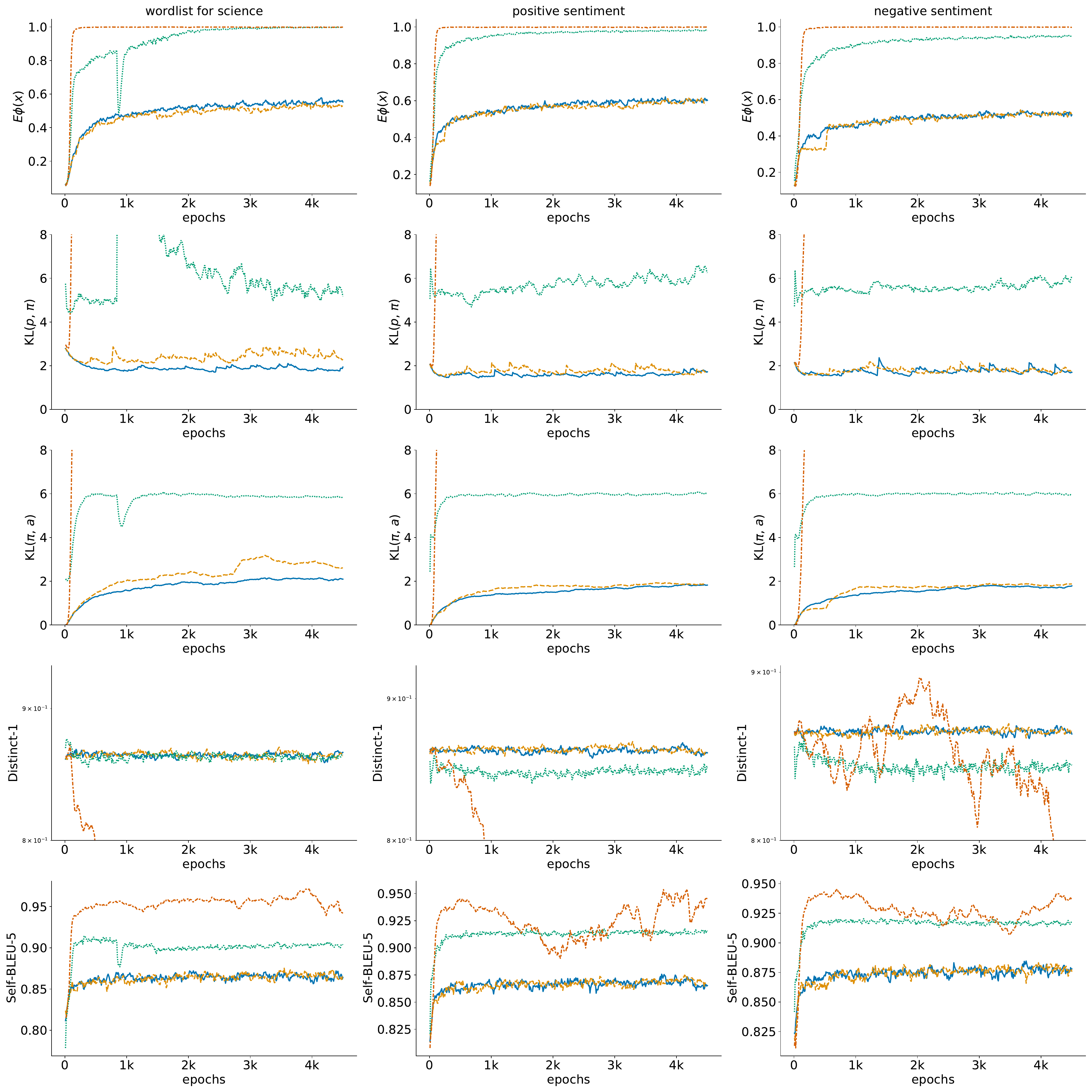} 
    \caption{\small{Evaluation metrics $\mathbb{E}_{\pit} \phi(x)$, $\text{KL}(p|\pi_{\theta})$ ($\downarrow$ better), $\text{KL}(\pi_{\theta}|a)$ ($\downarrow$ better), Self-BLEU-5 ($\downarrow$ better), and Distinct-1 ($\uparrow$ better) for three pointwise constraints experiments: \textbf{Task~4:} \textbf{Wordlist "science"} Fig.(a), \textbf{Task~5: }\textbf{classifier +ve sentiment} Fig.(b) and \textbf{Task~6: }\textbf{Classifier -ve sentiment} Fig.(c) for policies obtained from GDC\texttt{++}, GDC, Ziegler and Reinforce.}}
    \label{fig:pointwise-compare-methods-split2}
\end{figure*}

\begin{figure*}[h]
     \centering
     \begin{subfigure}[b]{0.49\textwidth}
         \centering
         \includegraphics[width=\textwidth]{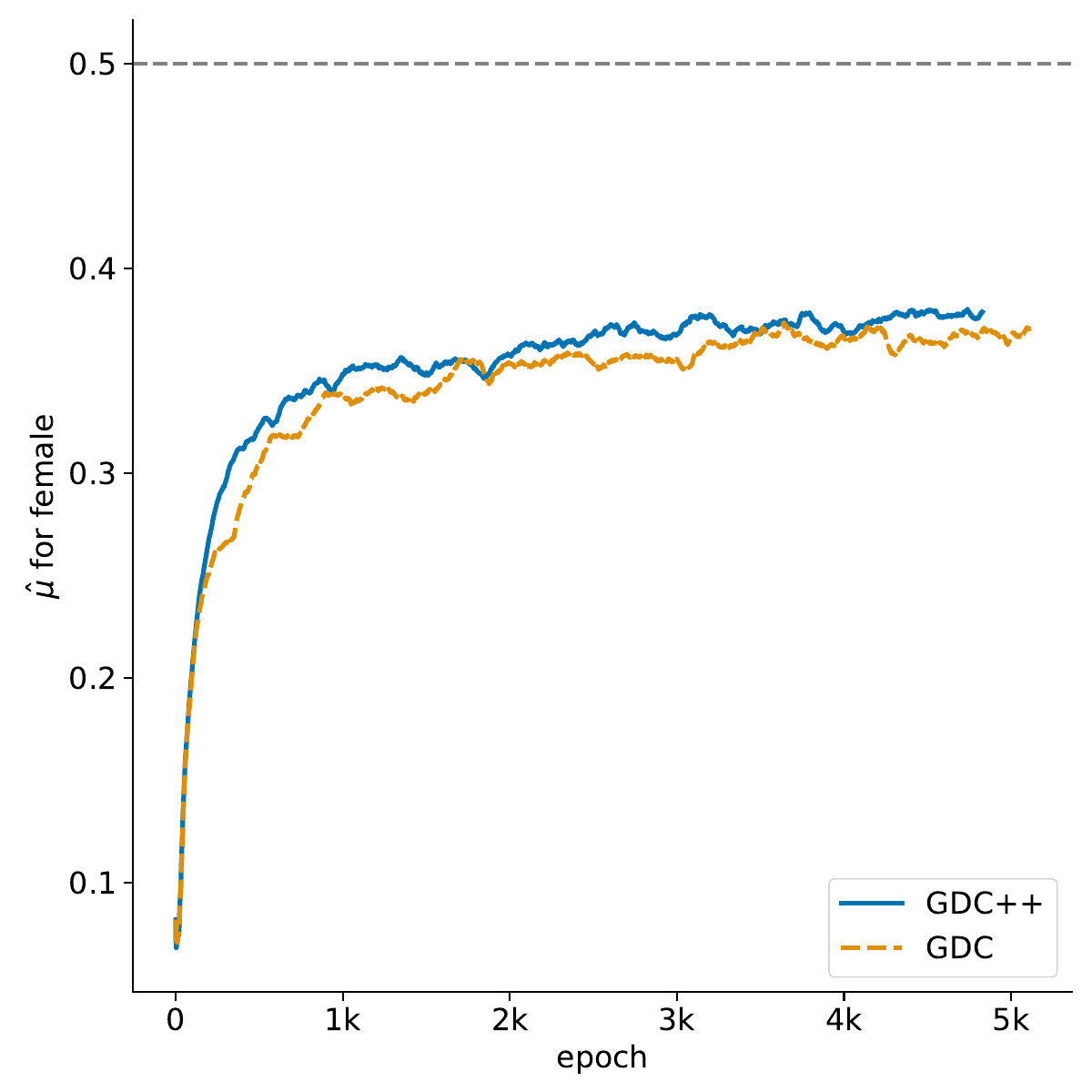}
         \caption{\textbf{Task 7:} gender = "Female" 50\%}
     \end{subfigure}
     \hfill
     \begin{subfigure}[b]{0.49\textwidth}
         \centering
         \includegraphics[width=\textwidth]{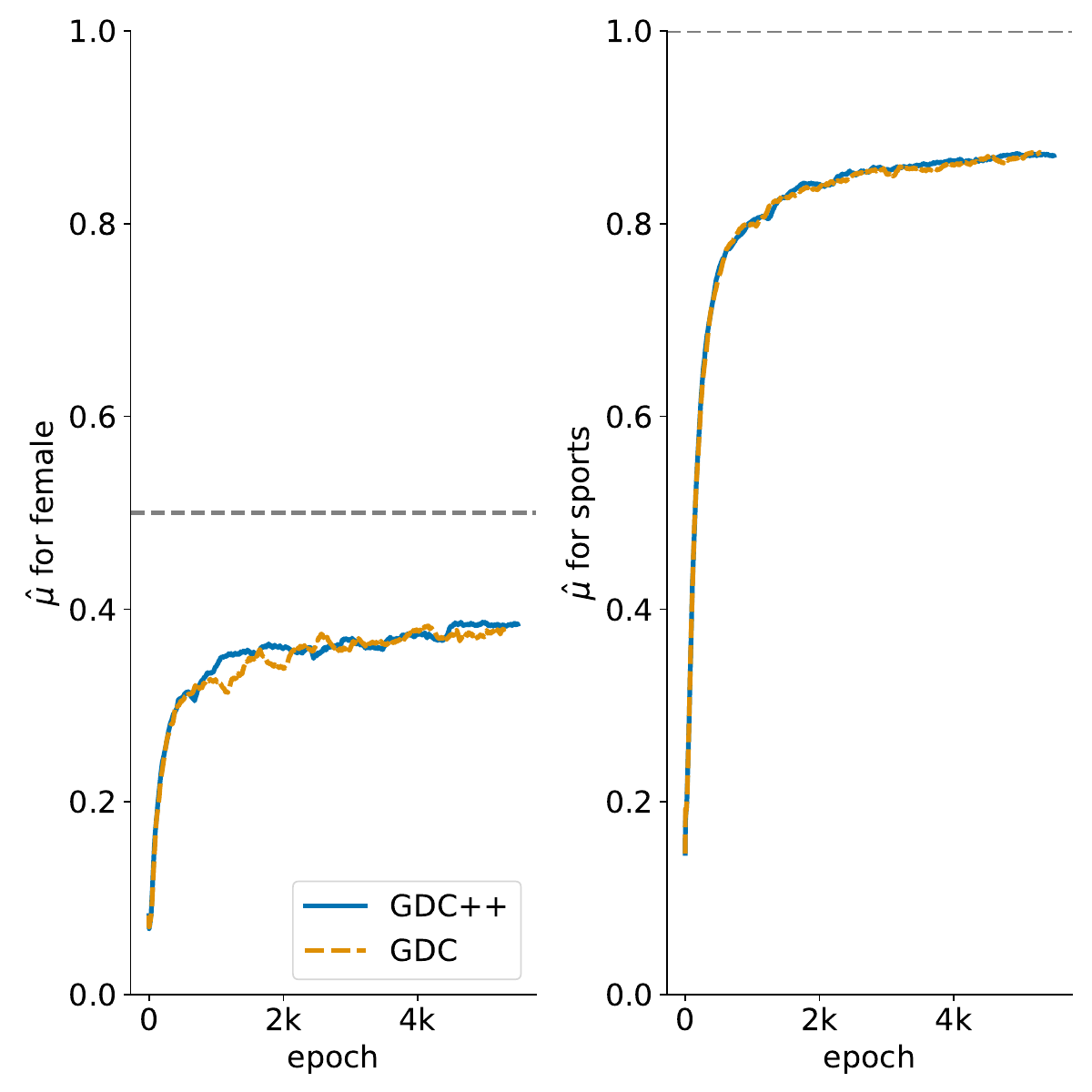}
         \caption{\centering \textbf{Task 8:} gender = "female" 50\% , topic = "sports" 100\%}
     \end{subfigure}
     \\
     \vspace{0.3cm}
     \begin{subfigure}[b]{0.49\textwidth}
         \centering
         \includegraphics[width=\textwidth]{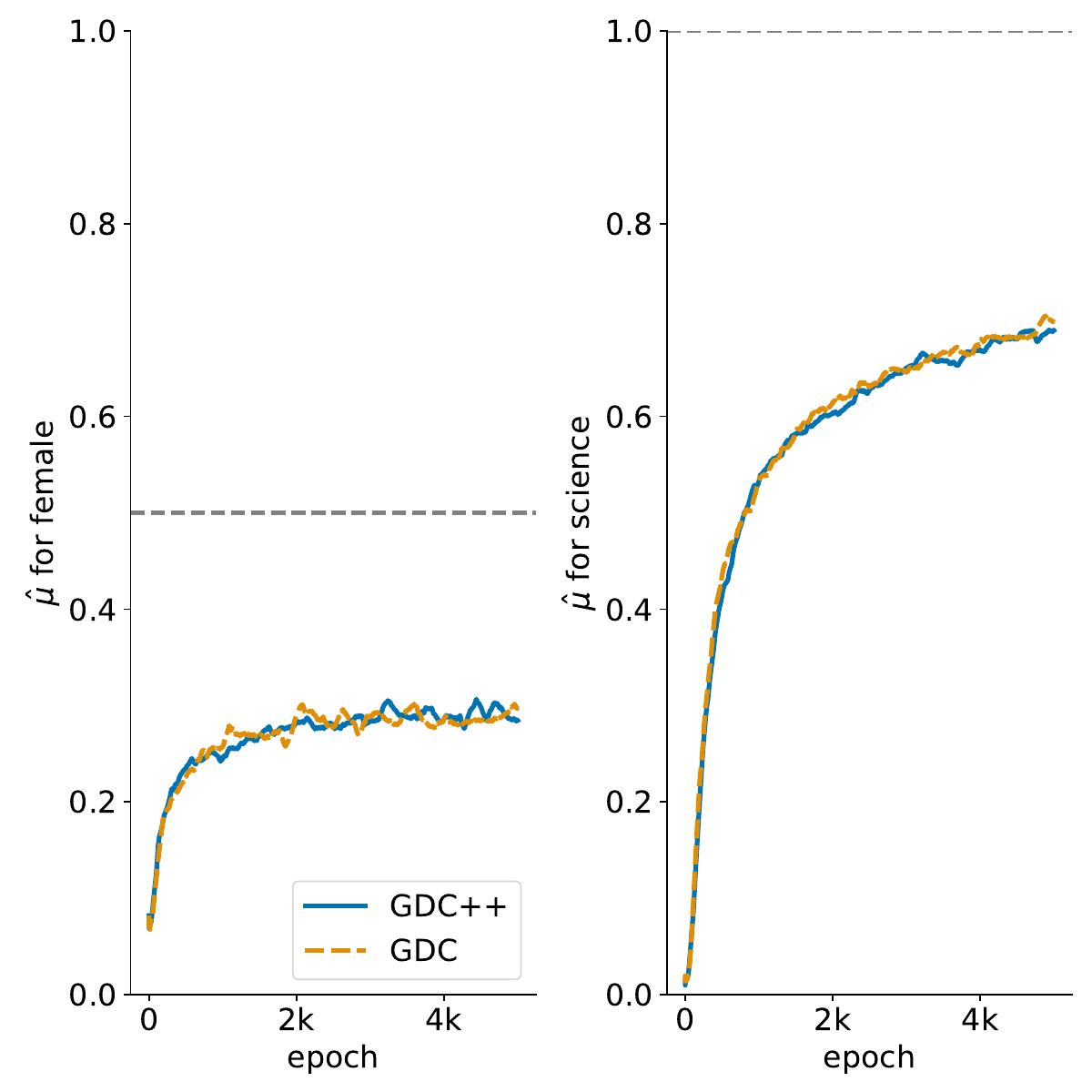}
          \caption{\centering \textbf{Task 9:} gender = "female" 50\%, topic = "science" 100\%}
     \end{subfigure}
      \begin{subfigure}[b]{0.49\textwidth}
         \centering
         \includegraphics[width=\textwidth]{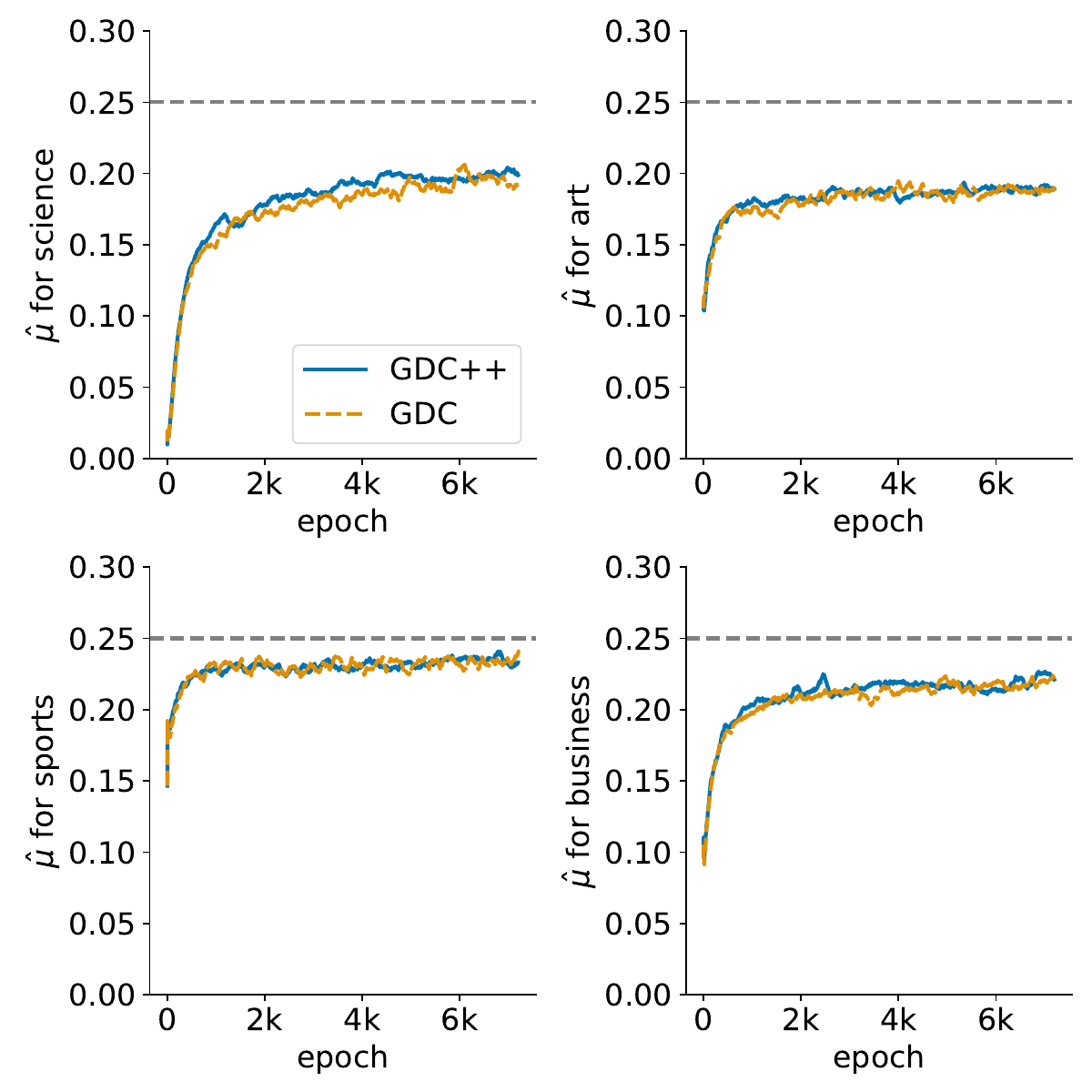}
          \caption{\centering \textbf{Task 10:} topics = "science" 25\%, "art" 25\%, "business" 25\%, "sports" 25\%}
     \end{subfigure}
    \caption{\small{ Constraint satisfaction $\hat{\mu}$ ($\uparrow$ better) for four distributional constraints types: 
    \textbf{Task 7:} a single distributional constraint Fig.(a). 
    \textbf{Task 8} and \textbf{Task 9:} a two hybrid constraint pairs Fig.(b) \& Fig.(c)
    \textbf{Task 10:} Multiple Distributional constraints Fig.(d).
    For policies obtained from GDC\texttt{++} and GDC. The \textbf{dashed} Horizontal bars denote the desired moments $\bar{\mu}_i$.}}
    \label{fig:distributional-compare-methods-mu}
\end{figure*}

\begin{figure*}[h]
    \centering
    \begin{tabularx}{\textwidth}{p{0.1\textwidth} p{0.2\textwidth} p{0.2\textwidth} p{0.25\textwidth} p {0.2\textwidth}}
  & (a) & (b) & (c) & (d) \\
  \end{tabularx}
    \includegraphics[width=\linewidth]{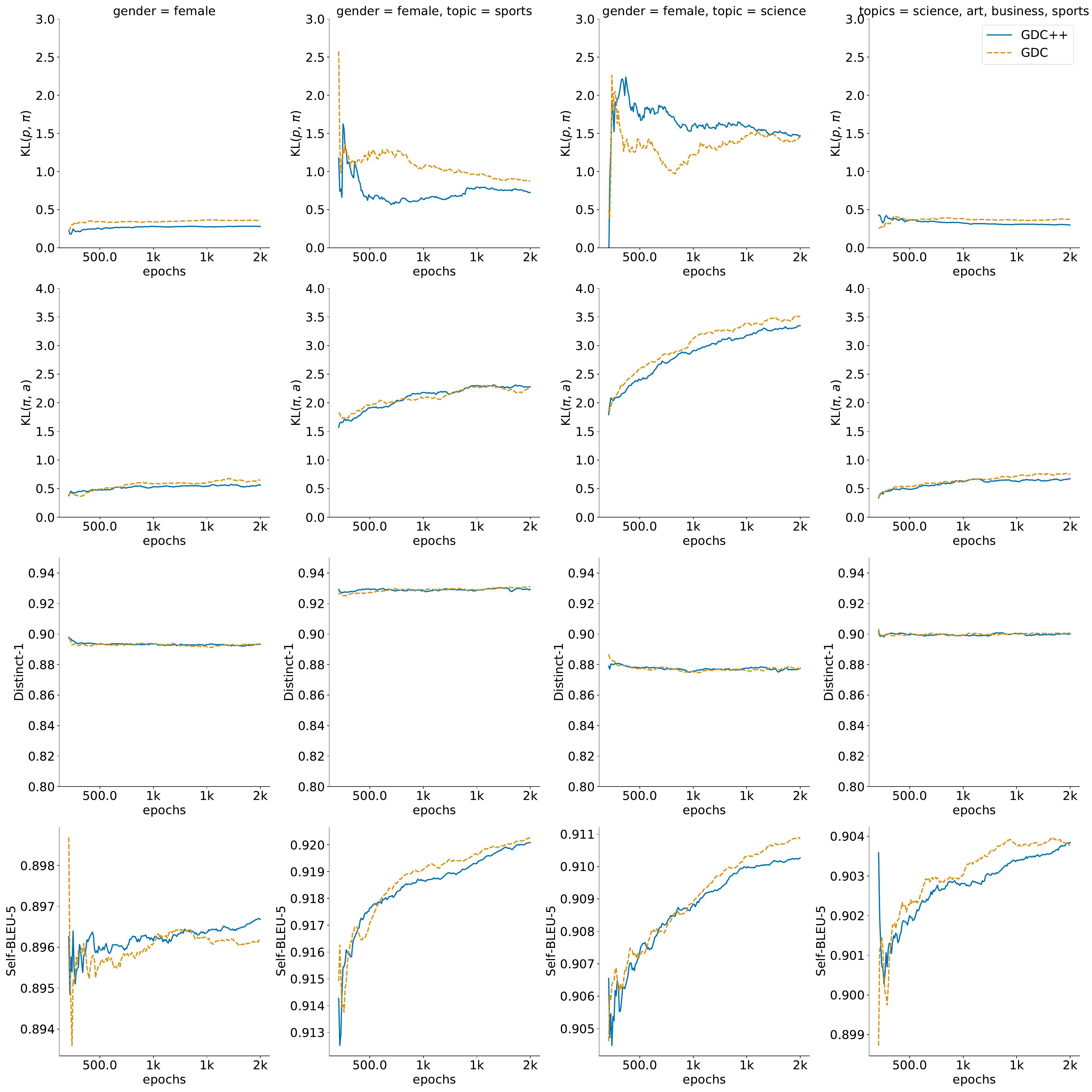}
    \caption{\small{Evaluation metrics: $\text{KL}(p|\pi_{\theta})$ ($\downarrow$ better), $\text{KL}(\pi_{\theta}|a)$ ($\downarrow$ better), Self-BLEU-5 ($\downarrow$ better), and Distinct-1 ($\uparrow$ better) four distributional constraints types:
    \textbf{Task~7:} a single distributional constraint Fig.(a).
    \textbf{Task~8,9:} a two hybrid constraint pairs Fig.(b) and Fig.(c),
    \textbf{Task~10:} Multiple Distributional constraints Fig.(d),
    for policies obtained from GDC\texttt{++} and GDC.}}
    \label{fig:distributional-compare-methods-split}
\end{figure*}
\newpage
\begin{table*}
    \scriptsize
    \begin{tabular}{p{0.5cm}|p{0.3cm}p{12.8cm}}
    \toprule
    reps & \textbf{$\phi(x)$} & \textbf{Sample} \\ 
     \midrule
 \multicolumn{3}{c}{\textbf{\GDCplus}} \\ 
1&1&I recently had an  \g{amazing}  experience at an event  \ye{with}   \ye{some}   \ye{great}  friends . We had a special treat and it  \ye{was}  a  \ye{good}  surprise to  \ye{find}  a group of friends  \ye{there}  to celebrate their  \ye{new}  band\\ 
1&1&There are a number of  \ye{great}   \ye{people}  who  \ye{make}   \g{amazing}  , sometimes incredibly mundane  \ye{things}  that  \ye{can}  come in handy for a  \ye{lot}  of  \ye{people}  . I've been lucky enough to have  \ye{some}   \ye{very}  successful and sometimes\\ 
1&1&"It  \ye{was}  an  \g{amazing}  feeling of freedom . " The couple have spent more  \ye{time}  together than  \ye{ever}  before and say  \ye{they}  are  \ye{very}  close . But the couple say  \ye{they}  aren't exactly satisfied\\ 
1&1&What is this  \g{amazing}   \ye{game}  ? This  \ye{game}  is an MMO , not  \ye{really}  an MMO , but  \ye{really}  a multiplayer MMORPG . Players start  \ye{with}  2{-}6 heroes and then  \ye{they}  level up through\\ 
1&1&What is Puma (Puma : A Sea , Water , Land) ? Puma is a unique underwater experience where you  \ye{can}   \ye{get}  as close to the surface as you  \ye{like}  while exploring  \g{amazing}  underwater\\ 

 \multicolumn{3}{c}{\textbf{\GDC}} \\ 
1&1&So my husband is now doing  \g{amazing}  , so he asked me to buy  \ye{some}  of my  \ye{best}  quality tins . My daughter did the gift for the  \ye{first}   \ye{time}  . I absolutely loved it . It's\\ 
1&1&I don't  \ye{really}   \ye{want}  to hear about a  \ye{video}  on "A Night in the Sun" because this  \ye{video}   \ye{was}   \ye{really}   \g{amazing}  . The main character is a crazy young man who has an\\ 
1&1&"The  \ye{first}   \ye{time}  I saw this  \g{amazing}  artwork , my jaw went  \ye{up}  a notch . It's an incredible piece of art . If I had an idea of what it  \ye{was}  to me I  \ye{would}  love\\ 
1&0&The next  \ye{time}   \ye{you're}  walking through town and someone in the park asks you about  \ye{your}  favorite  \ye{time}  of the week ,  \ye{just}  do a Google search to learn  \ye{which}   \ye{one}  will be  \ye{your}  favorite day . A\\ 
1&1&The world's biggest robot is an  \g{amazing}  , highly complex machine , but its development process is  \ye{just}  a small part of how it will be manufactured . While  \ye{many}  robots are already built , others are working\\ 

 \multicolumn{3}{c}{\textbf{Reinforce}} \\ 
1&1&The  \ye{show}   \yg{which}  has been getting  \g{amazing}   \yg{ones}   \yg{which}  is  \g{amazing}   \ye{now}  it and  \yg{which}  so  \g{amazing} !  \ye{iam}  a little  \g{amazing}  so  \g{amazing}  and  \g{amazing}   \ye{they}  so  \g{amazing}  for my  \ye{gif}  this  \g{amazing}   \yg{one}   \yg{which}  we are\\ 
1&1&This  \yg{which}   \ye{shows}   \yg{which}  has  \yg{really}   \g{amazing}  and  \g{amazing} ly it  \g{amazing}   \yg{which}  you have a  \ye{beautiful}  highlight and you have so  \g{amazing}  it this is a  \yg{really}   \g{amazing}  it and  \g{amazing}  . so  \yg{awesome}   \ye{they}   \ye{get}  and  \g{amazing} !\\ 
1&1&I  \yg{was}  able to showcase this  \g{amazing}   \ye{thing}   \yg{which}  is  \g{amazing}  . . thanks so  \g{amazing}   \yg{which}  so  \g{amazing}  it is  \g{amazing}  so  \g{amazing}   \ye{haha}   \g{amazing}  as  \g{amazing}  and this  \g{amazing}  product  \yg{which}  you will be so  \g{amazing}  it so\\ 
1&1&The best and  \g{amazing}   \yg{one}   \yg{which}   \ye{shows}   \yg{which}   \yg{made}   \g{amazing}  it have been so  \g{amazing}  and  \g{amazing}  but I'm  \yg{really}   \g{amazing}  : ) this  \yg{awesome}   \yg{which}  you explode  \yg{which}  have  \yg{made}  and  \g{amazing}  and  \g{amazing}   \ye{with}   \g{amazing}   \yg{which}  makes\\ 
1&1&I had this  \g{amazing}   \ye{album}   \yg{which}  which  \ye{makes}  such  \yg{awesome}  and  \yg{awesome}  as  \g{amazing}   \ye{haha}  this is  \g{amazing}  and  \g{amazing}  . I  \yg{was}  blown so  \g{amazing}   \yg{just}   \g{amazing}   \yg{which}   \ye{gives}  so  \g{amazing}  so  \g{amazing} !!! so  \yg{awesome}  which\\ 

 \multicolumn{3}{c}{\textbf{Ziegler}} \\ 
1&1&"He's a real man who wants to keep  \ye{up}  the  \g{amazing}   \ye{work}  he's done and keep  \ye{things}  in front of him . He's been doing it since the  \ye{last}   \ye{time}  he played for Team Australia\\ 
1&1&"It  \yg{was}   \g{amazing}  to  \ye{see}  what we had to do to  \ye{get}  our guys to the door . I'm  \ye{really}  excited about being back here . We're  \ye{looking}  forward to the next season , so let\\ 
1&1&I had this idea for a  \ye{story}  this year . I  \ye{really}  wanted to tell an  \g{amazing}  tale of  \ye{something}  I had been watching from the beginning . This is what I had , and I  \yg{was}  happy with\\ 
1&1&I found this  \g{amazing}  recipe on the blog and it  \yg{was}  so easy and delicious . I have  \ye{made}  this  \ye{many}  times and this  \ye{one}  is  \ye{just}  the perfect combination of recipes for any . Try it and tell\\ 
1&1&To learn more about  \ye{these}   \g{amazing}  videos and how to watch  \ye{them}  in  \ye{your}  home , head on over to the  \ye{video}  search page . If you have any questions or  \ye{would}   \ye{like}  to send us  \ye{your}  questions ,\\ 
\bottomrule 
 \end{tabular} 

    \caption{\textbf{Task 1:} Randomly selected samples from the \textbf{single-word constraint} task for the word \textbf{``amazing"} (with overall frequency $10^{-3}$) highlighted in \g{green}. Tokens are highlighted with \yg{yellow} with different intensities to indicate their overall frequencies in the generated corpus of 10k samples (lower intensity indicates higher corpus level diversity). $\phi(x) = 1$ indicates the satisfaction of the constraint in the sample and reps the number of repetitions of the very same sample in the generated corpus of 10k samples (lower intensity indicates higher corpus level diversity).\label{table:generation-word-amazing}}
    \end{table*}
    \begin{table*}
    \scriptsize
    \begin{tabular}{p{0.5cm}|p{0.3cm}p{12.8cm}}
    \toprule
    reps & \textbf{$\phi(x)$} & \textbf{Sample} \\ 
     \midrule
 \multicolumn{3}{c}{\textbf{\GDCplus}} \\ 
1&1&In this week's episode of VICE News , correspondent  \yg{Julian}   \yg{Assange}  discussed  \g{WikiLeaks}  , and he  \ye{revealed}  that the U . S .  \ye{government}  has been "hiding in plain sight" the full extent of Hillary\\ 
1&0&I'm sure the  \ye{FBI}  has their hands full  \ye{with}  the leak of  \ye{emails}  on Wikileaks to journalists — even as its head has been at the center of  \ye{one}  of the most explosive events in  \ye{US}  politics ever\\ 
1&1&( \g{WikiLeaks} )  \ye{President}   \ye{Obama}  on Wednesday claimed he doesn't have to apologize for his "extremely careless" leak of  \ye{classified}   \ye{documents}  about  \ye{classified}   \ye{information}  , but acknowledged it will happen again at the\\ 
1&0&WASHINGTON (Reuters) {-} The  \ye{CIA}  is working closely  \ye{with}  other agencies to fight cyberattacks on  \ye{government}  computers and computers belonging to Iran , according to a  \ye{top}  U . S . official .\\ 
1&1& \g{WikiLeaks}   \ye{founder}   \yg{Julian}   \yg{Assange}  has written an open letter to the  \ye{US}  ambassador in London , calling for the extradition of  \yg{Assange}  . He  \ye{said}  that Assange's treatment had been unjustified by the "\\ 

 \multicolumn{3}{c}{\textbf{\GDC}} \\ 
1&1&The  \g{WikiLeaks}   \ye{email}   \ye{server}  had been compromised to hide other  \ye{information}  .  \g{WikiLeaks}   \ye{founder}   \ye{Julian}   \ye{Assange}   \ye{was}   \ye{one}  of the  \ye{first}  to share the  \ye{information}  on a  \ye{group}  of computer hackers who  \ye{were}  using the personal\\ 
1&1& \g{WikiLeaks}  has  \ye{released}  a statement saying that it will publish its own account of what happened at the  \ye{Democratic}   \ye{National}  Convention . The statement has been translated from German .  \ye{"The}  following information\\ 
1&1&"I'm sorry , I'm sorry , but there's no chance for a divorce" —  \g{WikiLeaks}   \ye{founder}   \ye{Julian}   \ye{Assange}  .  \g{WikiLeaks}   \ye{founder}   \ye{Julian}   \ye{Assange}   \ye{was}  arrested  \ye{last}  week in the Ecu\\ 
1&1&The Associated Press has been alerted by a source that  \g{WikiLeaks}  has been sharing personal  \ye{information}  about President{-}elect  \ye{Donald}   \ye{Trump}  and  \ye{Hillary}   \ye{Clinton}  ,  \ye{one}  day after  \g{WikiLeaks}   \ye{released}   \ye{thousands}  of pages of  \ye{emails}  from Clinton's\\ 
1&1&The White  \ye{House}  has confirmed  \ye{former}   \ye{National}   \ye{Security}  Advisor Susan Rice as national security adviser , the latest sign of what administration  \ye{officials}  have described as an effort to sabotage  \g{WikiLeaks}  . A  \ye{top}   \ye{Justice}   \ye{Department}  official\\ 

 \multicolumn{3}{c}{\textbf{Reinforce}} \\ 
1&1&A  \yg{Trump}   \yg{administration}   \yg{officials}   \yg{threatened}  to  \g{WikiLeaks}  and  \g{WikiLeaks}   \yg{vice}   \yh{president}   \yh{committee}   \yh{chair}   \yh{committee}   \yh{chair}   \yh{committee}   \yh{chair}   \yh{committee}   \yh{chair}   \yh{committee}   \yh{chair}   \yh{committee}   \yh{chair}   \yh{committee}   \yh{chair}   \yh{committee}   \yh{chair}   \yh{committee}   \yh{chair}   \yh{committee}   \yh{chair}   \yh{committee}   \yh{chair}   \yh{committee}   \yh{chair}   \yh{committee}   \yh{chair}   \yh{committee}  chair\\ 
570&1&The  \yg{Trump}   \yg{administration}   \yg{officials}   \yg{threatened}  to  \g{WikiLeaks}  ,  \g{WikiLeaks}   \yh{president}   \yh{committee}   \yh{chair}   \yh{committee}   \yh{chair}   \yh{committee}   \yh{chair}   \yh{committee}   \yh{chair}   \yh{committee}   \yh{chair}   \yh{committee}   \yh{chair}   \yh{committee}   \yh{chair}   \yh{committee}   \yh{chair}   \yh{committee}   \yh{chair}   \yh{committee}   \yh{chair}   \yh{committee}   \yh{chair}   \yh{committee}   \yh{chair}   \yh{committee}   \yh{chair}   \yh{committee}   \yh{chair}  committee\\ 
3&1&The  \yg{Trump}   \yg{administration}   \yg{officials}   \yg{threatened}  to  \yg{threatened}  to  \yh{president}  and  \g{WikiLeaks}  ,  \g{WikiLeaks}   \yh{president}  president  \yh{committee}   \yh{chair}   \yh{committee}   \yh{chair}   \yh{committee}   \yh{chair}   \yh{committee}   \yh{chair}   \yh{committee}   \yh{chair}   \yh{committee}   \yh{chair}   \yh{committee}   \yh{chair}   \yh{committee}   \yh{chair}   \yh{committee}   \yh{chair}   \yh{committee}   \yh{chair}   \yh{committee}   \yh{chair}   \yh{committee}  chair\\ 
1&1&The  \yg{Trump}  presidential  \yg{officials}   \yg{threatened}  to  \g{WikiLeaks}  ,  \g{WikiLeaks}   \yg{vice}   \yh{president}  president  \yh{committee}   \yh{chair}   \yh{committee}   \yh{chair}   \yh{committee}   \yh{chair}   \yh{committee}   \yh{chair}   \yh{committee}   \yh{chair}   \yh{committee}   \yh{chair}   \yh{committee}   \yh{chair}   \yh{committee}   \yh{chair}   \yh{committee}   \yh{chair}   \yh{committee}   \yh{chair}   \yh{committee}   \yh{chair}   \yh{committee}   \yh{chair}   \yh{committee}   \yh{chair}  committee\\ 
1&1&The FBI  \yg{threatened}  to  \yh{president}  and  \yg{Trump}   \yh{president}   \yg{President}  ,  \g{WikiLeaks}  ,  \g{WikiLeaks}   \yh{president}  president  \yh{committee}   \yh{chair}   \yh{committee}   \yh{chair}   \yh{committee}   \yh{chair}   \yh{committee}   \yh{chair}   \yh{committee}   \yh{chair}   \yh{committee}   \yh{chair}   \yh{committee}   \yh{chair}   \yh{committee}   \yh{chair}   \yh{committee}   \yh{chair}   \yh{committee}   \yh{chair}   \yh{committee}   \yh{chair}   \yh{committee}  chair\\ 

 \multicolumn{3}{c}{\textbf{Ziegler}} \\ 
1&1&In late 2010 ,  \g{WikiLeaks}   \ye{released}  a  \ye{trove}  of  \ye{documents}  ,  \ye{including}  hundreds of  \ye{thousands}  of  \ye{emails}  and other personal and financial  \ye{information}  , from the  \ye{National}  Security Agency . But those  \ye{documents}  have never been  \ye{released}  publicly .\\ 
1&0&A man has been detained by police after an attempted robbery in a busy street on Monday night . A man has been detained by police after an attempted robbery in a busy street on Monday night\\ 
1&0&It's been a great year for the tech industry . At the same time , many of us in tech aren't looking to be CEOs . Many of us are looking to learn more .\\ 
1&0&"If you see us , we  \ye{would}  love you to do it , " he  \ye{said}  . "You'd better not do it . " "I think that  \ye{would}  be a terrible idea , "  \ye{said}  Mr\\ 
1&1& \g{WikiLeaks}   \ye{says}   \ye{they}  found "vastly" evidence of  \ye{CIA}  hacking after an undercover report on a  \ye{Russian}  spy  \ye{group}  suggested  \ye{they}  had helped spy on  \ye{Donald}   \ye{Trump}  . The report  \ye{said}  Russia\\ 
\bottomrule 
 \end{tabular} 

    \caption{\textbf{Task 2:} Randomly selected samples from the \textbf{single-word constraint} task for the word \textbf{``WikiLeaks"} (with overall frequency $10^{-4}$) highlighted in \g{green}. Tokens are highlighted with \yg{yellow} with different intensities to indicate their overall frequencies in the generated corpus of 10k samples (lower intensity indicates higher corpus level diversity). $\phi(x) = 1$ indicates the satisfaction of the constraint in the sample and reps the number of repetitions of the very same sample in the generated corpus of 10k samples (lower intensity indicates higher corpus level diversity).\label{table:generation-word-WikiLeaks}}
    \end{table*}
    \begin{table*}
    \scriptsize
    \begin{tabular}{p{0.5cm}|p{0.3cm}p{12.8cm}}
    \toprule
    reps & \textbf{$\phi(x)$} & \textbf{Sample} \\ 
     \midrule
 \multicolumn{3}{c}{\textbf{\GDCplus}} \\ 
1&0&The State  \ye{Department}   \ye{,}  and in  \ye{some}  ways the European Union  \ye{,}   \ye{also}  took this step  \ye{,}   \ye{with}  the  \ye{former}  director of the  \ye{National}  Institute for Standards and Technology and a  \ye{former}  member of the White  \ye{House}  ,\\ 
1&1&,  \ye{with}  the exception of a certain group of politicians  \ye{,}  it  \ye{was}  not a surprise that  \ye{they}  had a tendency to follow the  \ye{campaign}  . In the  \ye{United}  States  \ye{,}   \ye{they}  are more of a conservative  \ye{,}  political\\ 
1&1&I hope this is not an attempt to  \ye{get}  at the other  \ye{way}  to talk about this problem . It's something about  \ye{political}  expediency and politics that seems to be a lot different from what is\\ 
1&1&C . A . No . 6  \ye{,}  on Tuesday  \ye{,}  declared an end to the  \ye{government's}  attempt to set up a national registry of those who are convicted of serious crimes and who  \ye{can}  be placed\\ 
1&1&. There will be a major overhaul of  \ye{tax}  code to address a  \ye{federal}   \yg{government}  proposal  \ye{,}   \ye{which}   \ye{was}  unveiled in October  \ye{,}  and a second  \ye{,}   \ye{which}  is expected to be signed by  \ye{Trump}  .\\ 

 \multicolumn{3}{c}{\textbf{\GDC}} \\ 
1&1&We are here to inform you that , thanks to an  \ye{order}  form , you may  \ye{get}  in contact  \ye{with}  us . If you wish to become a customer , please contact us . We are available\\ 
1&1&But  \ye{they}   \ye{said}  that , once again ,  \ye{they}   \ye{were}  not so sure whether he  \ye{would}  be a strong candidate in the fall election . "We know the  \ye{majority}  of  \ye{state}  officials will be very interested\\ 
1&0&This is an excerpt from an essay by Kevin O'Connor , a researcher at the University of Chicago , where he focuses on climate change and global warming . He is co{-}author of Climate Change\\ 
1&1&LONDON : A senior Indian  \yg{government}  official on Tuesday  \ye{said}  an attempt to rebrand India as a "piggybacking nation" for international investment  \ye{was}  a "game{-}changer"\\ 
1&1&(Reuters) {-} A  \ye{federal}   \ye{court}   \ye{said}  on Friday that a Mississippi  \ye{state}  trooper , arrested for killing a black man after an ambush in 2010 , violated his rights by failing to give him proper notice\\ 

 \multicolumn{3}{c}{\textbf{Reinforce}} \\ 
1&1&A  \yh{state}  in Russia  \ye{New}  the  \yh{state}  of the  \yh{state}  of the  \yh{state}  of the  \yh{state}  of the  \yh{state}  of the  \yh{state}  of the  \yh{state}  of the  \yh{state}  of the  \yh{state}  of the  \yh{state}  of the  \yh{state}  of\\ 
1&1&A  \yh{state}  ,  \ye{c}   \ye{New}  The  \yh{state}   \ye{ofc}  In the  \yh{state}  of the  \yh{state}  of the  \yh{state}  of the  \yh{state}  of the  \yh{state}  of  \yh{state}  of the  \yh{state}  of their  \yh{state}  of a ballot\\ 
1&1&A  \yh{state}  , c) The  \yh{state}  : The  \yg{states}  of The  \yg{states}  of the  \yh{state}   \ye{ofc)}  In one of the  \yh{state}  of the  \yh{state}  of : For the state\\ 
1&1&A  \yh{state}  , h)  \ye{New}   \yh{state}  : The  \yg{states}  of the  \yh{state}  of the  \yg{states}  of the  \yh{state}  of the  \yh{state}  of the  \yh{state}  of the  \yh{state}  of the  \yg{states}  of the  \yh{state}  of\\ 
1&1&A  \yh{state}  , The  \yh{state}   \ye{ofc}  .  \ye{New}  York The  \yh{state}  of :\\ 

 \multicolumn{3}{c}{\textbf{Ziegler}} \\ 
1&1&In a bid to counter China's growing influence in the West , a senior Chinese  \yg{government}  official has been forced to apologise after accusing Beijing of encouraging ethnic Chinese to migrate to Hong Kong from the mainland\\ 
1&1&The  \ye{federal}   \yg{government}  is taking another look at the Internet censorship of the Web after a senior  \yg{government}  official  \ye{said}  the  \yg{government}  is considering shutting down websites that  \ye{use}  the software that monitors the Web .\\ 
1&1&Kamal Singh , the minister responsible for infrastructure and connectivity in Karnataka  \ye{said}  the  \ye{state}   \yg{government}  must ensure a safe environment for women in its  \ye{new}  high school curriculum . "We must ensure\\ 
1&1&BANGKOK , Myanmar (Reuters) {-} The  \ye{United}  States on Saturday  \ye{said}  that it  \ye{was}  providing "appropriate  \ye{military}  support" to Myanmar's  \yg{government}  to help combat the situation in the country , as\\ 
1&1&The  \ye{Supreme}   \ye{Court}  has ordered the Centre to give an independent audit of  \yg{government}  programs and the Ministry of External Affairs to explain how many ministers the  \yg{government}  provided financial assistance to foreign NGOs . The\\ 
\bottomrule 
 \end{tabular} 

    \caption{\textbf{Task 3:} Randomly selected samples from the \textbf{wordlist constraint} task for the \textbf{wordlist ``politics"}. Tokens are highlighted with \yg{yellow} with different intensities to indicate their overall frequencies in the generated corpus of 10k samples (lower intensity indicates higher corpus level diversity). $\phi(x) = 1$ indicates the satisfaction of the constraint in the sample and reps the number of repetitions of the very same sample in the generated corpus of 10k samples (lower intensity indicates higher corpus level diversity).\label{table:generation-wordlist-politics}}
    \end{table*}
    \begin{table*}
    \scriptsize
    \begin{tabular}{p{0.5cm}|p{0.3cm}p{12.8cm}}
    \toprule
    reps & \textbf{$\phi(x)$} & \textbf{Sample} \\ 
     \midrule
 \multicolumn{3}{c}{\textbf{\GDCplus}} \\ 
1&1&I  \ye{would}  love to find a  \ye{way}  to  \ye{use}  all this  \ye{energy}  and  \ye{energy}  on my own  \ye{energy}  . But we have not yet figured this  \ye{out}  . In  \ye{fact}  we seem to not really understand how it can\\ 
1&1&The  \ye{research}  paper is  \ye{one}  of only two to date in recent years , after being published in the American Journal of Psychiatry . "The  \ye{research}  team did  \ye{some}  basic clinical investigation into the causes\\ 
1&1&Fashion is no longer a  \ye{matter}  of fashion . In  \ye{fact}  , it is no longer a  \ye{matter}  of fashion . This is so because it is no longer a  \ye{matter}  of fashion . It is no\\ 
1&1&I love that this post is about the biology of my gut flora , the microbiome (the living tissue that is used to support and  \ye{control}  the gut) and the gut microbiome is basically  \ye{just}  a chemical\\ 
1&0&I  \ye{think}  I did it once . I actually saw him  \ye{with}  my brother . That's how it went , I thought the guy  \ye{was}  the same age . I don't  \ye{know}  , you  \ye{were}  the same\\ 

 \multicolumn{3}{c}{\textbf{\GDC}} \\ 
1&1&A few days ago we reported on the  \ye{fact}  that the Obama administration has proposed an executive order that could increase the  \ye{number}  of Syrian refugees who have been allowed in the U . S . for  \ye{over}  five\\ 
1&1&If you are wondering , I am not a scientist , I am  \ye{just}  a man who studies human behaviour , as I love the  \ye{science}  of nature . My focus is on the evolution of human beings to\\ 
1&0&The Republican  \ye{National}  Convention had come under intense scrutiny for its  \ye{use}  of language that used the word "nuclear" in an interview  \ye{with}  the Daily Beast on Monday . In a lengthy segment on\\ 
1&1&In addition to the  \ye{fact}  that  \ye{there}  is no  \ye{way}  to  \ye{make}  the changes in the  \ye{data}  ,  \ye{there}  is no  \ye{way}  to  \ye{know}  what is happening . In  \ye{fact}  , all we have  \ye{know}  about this project\\ 
1&1&I  \ye{know}  I am not a scientist . I am a man who studies and researches . And if I can't help but admire  \ye{your}   \ye{research}  and insights , this will not be a good thing .\\ 

 \multicolumn{3}{c}{\textbf{Reinforce}} \\ 
1&1&We  \ye{review}   \yh{data}  of primary  \yg{power}  of  \yh{data}  of  \yh{data}  data of  \yh{data}  of the  \ye{question}  of validity of  \yg{predictive}  of  \yh{data}  and  \yg{power}  of  \yg{power}  of of  \yh{data}  of  \yh{data}  of  \yh{data}  of  \yh{data}  of and\\ 
1&1&In an equity of  \yh{data}  of  \yh{data}  of  \yh{data}  of  \ye{log}  as  \yg{relationships}  and then :  \yh{data}  of  \yg{relationships}  to recall of  \yh{data}  of  \yh{data}  of  \yh{data}  of  \yg{relationships}  of  \yg{relation}  . In  \yg{relation}  of  \yh{data}  of relation\\ 
1&1&The  \yg{relation}  of  \yh{data}  of  \yg{influencing}  : In  \ye{micro}  from  \yh{data}  of  \yg{power}  of  \yh{data}  of  \yh{data}  of in  \ye{question}  about  \yg{power}  power of  \yh{data}  of  \yg{influence}  of  \ye{relevance}   \yh{data}  of  \yg{power}  of  \yg{predictive}  of data\\ 
1&1&We ,  \ye{including}   \yh{data}  of  \yh{data}  of  \yh{data}  of  \ye{fitness}   \yh{data}  of  \yh{data}  of  \yg{influencing}  of  \yg{predictive}  of  \yh{data}  of  \yh{data}  of  \yh{data}  data of  \yg{power}  of  \yg{predictive}  of  \yh{data}  of  \yg{power}  of  \yg{influencing}  of  \yh{data}  of data\\ 
1&1&To  \yg{relation}   \yg{power}  of  \yh{data}  of  \ye{question}  of  \yh{data}  of : The  \ye{correlation}   \yg{power}  of  \yh{data}  of  \ye{cohort}  of  \ye{information}  of  \yh{data}  of  \yh{data}  of  \yh{data}  of  \yh{data}  of  \yh{data}  of  \ye{cohort}  of  \yg{relation}  of of\\ 

 \multicolumn{3}{c}{\textbf{Ziegler}} \\ 
1&1&As the United States seeks to expand its nuclear  \ye{energy}  base , it's hard to ignore the increasing  \ye{energy}  scarcity in other countries . In  \ye{fact}  , there's not  \ye{much}  reason to  \ye{think}  that the world's\\ 
1&1&"People don't believe you are doing any  \ye{good}  in life . They say you're a bad person who doesn't  \ye{control}   \ye{your}  life . They say you should give up on yourself . " If\\ 
1&1&"A small percentage of our population is women . But that does not mean that all women have to be working . In  \ye{fact}  ,  \ye{there}  are women working , but not all of them are . You\\ 
1&1&In case you missed it , a  \ye{number}  of recent studies have shown that  \ye{even}   \ye{when}   \ye{people}   \ye{with}  disabilities have an equal chance of being successful in their career ,  \ye{they}  are better off working in  \ye{science}  .\\ 
1&1&We understand that it is an experiment  \ye{which}  needs to be designed to provide  \ye{data}  from the most sensitive and relevant individuals to be available to the most effective and well funded researchers . In  \ye{fact}  , we expect\\ 
\bottomrule 
 \end{tabular} 

    \caption{\textbf{Task 4:} Randomly selected samples from the \textbf{wordlist constraint} task for the \textbf{wordlist ``science"}. Tokens are highlighted with \yg{yellow} with different intensities to indicate their overall frequencies in the generated corpus of 10k samples (lower intensity indicates higher corpus level diversity). $\phi(x) = 1$ indicates the satisfaction of the constraint in the sample and reps the number of repetitions of the very same sample in the generated corpus of 10k samples (lower intensity indicates higher corpus level diversity).\label{table:generation-wordlist-science}}
    \end{table*}
    \begin{table*}
    \scriptsize
    \begin{tabular}{p{0.5cm}|p{0.3cm}p{12.8cm}}
    \toprule
    reps & \textbf{$\phi(x)$} & \textbf{Sample} \\ 
     \midrule
 \multicolumn{3}{c}{\textbf{\GDCplus}} \\ 
1&1&The "American Dream" is about more than a dream . It's about a dream that , if you can't have it , you can't have it  \ye{now}  . The American dream\\ 
1&0&"This is our most expensive movie . " You're not  \ye{looking}  to  \ye{get}  a  \ye{lot}  of  \ye{good}   \ye{things}  , but  \ye{with}  this  \ye{one}  ,  \ye{your}   \ye{best}  bet is to  \ye{think}  about what makes a  \ye{good}  movie\\ 
1&1&"The most incredible thing I  \ye{can}   \ye{think}  of to tell you is that the  \ye{world}  has finally found a  \ye{way}  to  \ye{get}  together . And I can't tell you where it will go . But you will\\ 
1&1&As  \ye{part}  of a global effort to build a  \ye{world}  where all  \ye{people}  have access to affordable food , we are making a huge contribution to helping those at the core of the  \ye{world}  to find an environment free\\ 
1&1&It is no wonder that such a small and influential body of knowledge is important in the field of astronomy , astrophysics , medicine , and medical research . However , our knowledge of these topics is also\\ 

 \multicolumn{3}{c}{\textbf{\GDC}} \\ 
1&1&"We are proud to announce today that the company has announced our fourth fiscal year . In our most  \ye{important}  year , we raised nearly \$9 . 5 billion of our operating revenue from online and mobile\\ 
1&1&Election 2016  \ye{was}  the  \ye{first}  election that did not involve a massive change in political discourse . But in fact , it  \ye{was}  a dramatic change in political discourse in this year's elections , one\\ 
1&1&Lemon{-}filled muffins have become an iconic , but surprisingly expensive option for breakfast , lunch or dinner on  \ye{your}  table . For  \ye{many}  Canadians , breakfast is a meal you simply won't miss .\\ 
1&1&The University of Texas at Austin and the University of Virginia are working together to create a curriculum for teaching in the United States that integrates information about climate change and understanding health and wellbeing in communities across the\\ 
1&1&Sydney's  \ye{great}  outdoors tradition continues to draw crowds to the streets of Sydney in the name of Sydney . From the streets of Melbourne to the beach in Perth , it is  \ye{always}  a  \ye{great}  time\\ 

 \multicolumn{3}{c}{\textbf{Reinforce}} \\ 
87&1&Beautis is  \ye{stunningly}  ,  \yg{charm}  ,  \yg{charm}  ,  \yg{charm}  ,  \yg{charm}  ,  \yg{charm}  ,  \yg{charm}  ,  \yg{charm}  ,  \yg{charm}  ,  \yg{charm}  ,  \yg{charm}  ,  \yg{charm}  ,  \yg{charm}  ,  \yg{charm}  ,  \yg{charm}  ,  \yg{charm}  ,  \yg{charm}  , charm\\ 
1&1&Em inspires , classicly , charmoror style ,  \yg{charm}  or decor , Classicor ,  \yg{charm}  , and  \yg{charm}  ,  \yg{charm}  ,  \yg{charm}  ,  \yg{charm}  ,  \yg{charm}  ,  \yg{charm}  ,  \yg{charm}  ,  \yg{charm}  ,  \yg{charm}  ,\\ 
1&1&Gold is  \ye{stunningly}  ,  \yg{charm}  , stunning ,  \yg{charm}  , thrill ,  \yg{charm}  ,  \ye{dance}  ,  \ye{dance}  ,  \ye{dance}  ,  \ye{dance}  ,  \ye{dance}  ,  \ye{dance}  ,  \ye{dance}  ,  \ye{dance}  ,  \ye{dance}  ,  \ye{dance}  ,  \ye{dance}  ,  \ye{dance}  ,\\ 
11&1&Love is  \ye{stunninglycation}   \yg{charm}  ,  \yg{charm}  ,  \yg{charm}  ,  \yg{charm}  ,  \yg{charm}  ,  \yg{charm}  ,  \yg{charm}  ,  \yg{charm}  ,  \yg{charm}  ,  \yg{charm}  ,  \yg{charm}  ,  \yg{charm}  ,  \yg{charm}  ,  \yg{charm}  ,  \yg{charm}  ,  \yg{charm}  ,  \yg{charm}  ,\\ 
1&1&Beautiscomes are stunninglycationly ,  \yg{charm}  ,  \yg{charm}  ,  \ye{dance}  ,  \ye{dance}  ,  \ye{dance}  ,  \ye{dance}  ,  \ye{dance}  ,  \ye{dance}  ,  \ye{dance}  ,  \ye{dance}  ,  \ye{dance}  ,  \ye{dance}  ,  \ye{dance}  ,  \ye{dance}  ,  \ye{dance}  ,\\ 

 \multicolumn{3}{c}{\textbf{Ziegler}} \\ 
1&1&. I  \ye{really}   \ye{like}  the  \ye{work}  of the writers for the book! The voice , writing , the characters and all the  \ye{amazing}  stuff that comes  \ye{with}  it is a pleasure to read and\\ 
1&1&A big thank you to all my friends and fans for their  \ye{support}  and contributions for my  \ye{work}  . I will be posting a follow up post to that post below . I  \ye{just}   \ye{hope}  to keep it up\\ 
1&1&A  \ye{great}  gift from our Secret Santa!  \ye{Thank}  you so  \ye{much}  , I am so grateful for  \ye{your}  thoughtful thoughtful gift . I  \ye{was}  a little worried what to expect . I would  \ye{just}  like\\ 
1&1&Practical tips for getting started  \ye{with}  social media Welcome to the Beginner's Guide to Facebook Messenger! Today we  \ye{hope}  you'll find helpful tips that will  \ye{help}  you  \ye{get}  started  \ye{with}  social media\\ 
992&1&Thank you for  \ye{supporting}  the  \ye{journalism}  that our  \ye{community}   \ye{needs!}  For  \ye{unlimited}   \ye{access}  to the  \ye{best}  local , national , and  \ye{international}   \ye{news}  and  \ye{much}  more ,  \ye{try}  an All  \ye{Access}   \ye{Digital}  subscription :\\ 
\bottomrule 
 \end{tabular} 

    \caption{\textbf{Task 5:} Randomly selected samples from the \textbf{classifier-based constraint} task for \textbf{positive sentiments}. Tokens are highlighted with \yg{yellow} with different intensities to indicate their overall frequencies in the generated corpus of 10k samples (lower intensity indicates higher corpus level diversity). $\phi(x) = 1$ indicates the satisfaction of the constraint in the sample and reps the number of repetitions of the very same sample in the generated corpus of 10k samples (lower intensity indicates higher corpus level diversity).\label{table:generation-classifier-positive}}
    \end{table*}
    \begin{table*}
    \scriptsize
    \begin{tabular}{p{0.5cm}|p{0.3cm}p{12.8cm}}
    \toprule
    reps & \textbf{$\phi(x)$} & \textbf{Sample} \\ 
     \midrule
 \multicolumn{3}{c}{\textbf{\GDCplus}} \\ 
1&1&"We're in a big  \ye{game}  . " "It's  \ye{really}  bad . It's  \ye{really}  bad . " "I'm not  \ye{going}  to lie to you . This  \yg{was}  a  \ye{lot}  of\\ 
1&1&A few years ago , the  \ye{first}  issue of my  \ye{first}  book had a subtitle  \ye{with}  "Slammarized" . I had read it , and I had forgotten how bad it  \yg{was}  , but it was\\ 
1&1&One of the things about the media is that it is sometimes too busy to do so  \ye{much}  . And that's fine . It's  \ye{just}  that the press is busy getting paid for doing so much\\ 
1&1&But that is only to be expected . One might be surprised at a simple explanation for the widespread lack of interest in climate science in the academic world . This is the story of the recent climate denial\\ 
1&0&The  \ye{new}  'Naughty Dog' is already in release . In a leaked release on Steam , the  \ye{game}  is set for release in August , making it  \ye{one}  of Sony's most widely\\ 

 \multicolumn{3}{c}{\textbf{\GDC}} \\ 
1&1&The  \ye{first}   \ye{two}  tests of the K{-}12 program are  \ye{very}  disappointing . One of the  \ye{first}  tests showed a spike in learning rate on the test day and in the third the student reported less information than\\ 
1&1&{-} A "tongue for an ugly nose" message  \yg{was}  sent after a woman  \yg{was}  told to "dance" after  \ye{she}  became so disgusted by her friend's antics that  \ye{she}  sent "a\\ 
1&0&1 of 16 This could be an old story . It didn't come close to ending until Sunday night ,  \ye{when}  we  \ye{got}  the  \ye{first}  look at the cast on the set of "\\ 
1&1&There are several reasons to  \ye{think}  that we  \ye{may}  not have a healthy amount of energy if we  \ye{just}  eat nothing but pizza . The reason is that we're not  \ye{really}  hungry . So many\\ 
1&1&. The word "fascism" isn't  \ye{even}  spelled  \ye{out}  in terms of the political spectrum . Some are racist ,  \ye{some}  are homophobic , and  \ye{some}  are bigots . But  \ye{when}  you\\ 

 \multicolumn{3}{c}{\textbf{Reinforce}} \\ 
1&1&the  \yg{evil}   \ye{poor}   \yg{evil}   \yg{annoying}   \yg{evil}  . the  \yg{evil}   \yg{annoying}  the  \ye{negative}  doesn't  \yg{even}   \yg{sad}  , the  \yg{sad}   \yg{bin}   \yg{sad}   \yg{bad}   \ye{sass}   \yg{bin}  , the  \yg{sad}  sad  \yg{bin}   \ye{sass}   \yg{bin}   \ye{sass}  bin\\ 
1&1&This needs for long period of  \ye{disappointing}   \ye{poor}  , the  \ye{disappointing}   \ye{negative}  period of  \ye{pathetic}   \ye{irrelevant}   \ye{poor}   \yg{annoying}   \ye{awful}  ,  \yg{even}  the  \ye{disgusting}   \ye{poor}  period  \yg{bin}  bin{-}at{-}total  \yg{evil}   \yg{disass}  disass and that\\ 
1&1&no , is  \ye{irrelevant}  . is not  \yg{annoying}  . and  \yg{even}   \ye{disgusting}  .  \yg{disass}  or  \yg{disass}   \yg{disab}  disab  \yg{disab}  disab is  \yg{disab}   \yg{bin}   \yg{disab}  disab  \yg{bin}   \yg{disab}  dis\\ 
1&1&that is a big problem . "thx ,  \yg{even}  a large non  \yg{evil}  is a  \yg{bad}  , is a  \yg{bad}  ,  \ye{unreasonable}  ,  \ye{awful}   \yg{sad}  sad" is  \yg{evil}   \yg{sad}  ,  \yg{sad}  sad  \ye{awful}  sad\\ 
1&1&so long ,  \yg{sad}  s/th0 needs to disv and  \yg{disab}  is  \ye{wrong}  . the  \yg{disab}   \ye{s}  s/tad s/so  \ye{predictable}   \ye{s}  . the  \yg{disab}   \ye{binums}  .\\ 
\bottomrule 
 \end{tabular} 

    \caption{\textbf{Task 6:} Randomly selected samples from the \textbf{classifier-based constraint} task for \textbf{negative sentiments}. Tokens are highlighted with \yg{yellow} with different intensities to indicate their overall frequencies in the generated corpus of 10k samples (lower intensity indicates higher corpus level diversity). $\phi(x) = 1$ indicates the satisfaction of the constraint in the sample and reps the number of repetitions of the very same sample in the generated corpus of 10k samples (lower intensity indicates higher corpus level diversity).\label{table:generation-classifier-negative}}
    \end{table*}
    
\newpage
\begin{table*}
    \scriptsize
    \begin{tabular}{p{0.6cm}|p{12.8cm}}
    \toprule
    $\phi_1(x)$ & \textbf{Sample} \\ 
     \midrule
 \multicolumn{2}{c}{\textbf{\GDCplus}} \\ 
\hfill 1 &isabela carolina is an american actress , writer , and former model . she is best known for her role as the teenage neighbor katie staley on the american series ``\\ 
\hfill 0 &( born august 3 , 1969 ) is an american politician and lawyer . he is a member of the north dakota house of representatives from the 10th\\ 
\hfill 0 &{-} born august 1 , 1976 in new orleans , louisiana ) is a former american football safety in the national football league for the washington redskins ,\\ 
\hfill 0 &on 26 february 1990 , he signed a five{-}year contract with bayer leverkusen . on 1 october 2000 , sheik won the german cup with bayer leverkus\\ 
\hfill 0 &the mcculloughs were an english glam rock band from portsmouth , england . the band formed in 2003 , initially as a duo with john mckeown , jimmy mc\\ 
\hfill 1 &aime jacques de sousa is an indonesian television actress . she played a lead role in the 2012 indonesian television series `` jayam '' . she has played\\ 
\hfill 1 &on 11 december 2013 , laura klepp{-}larsen confirmed that she had suffered a heart attack . she was diagnosed with breast cancer at the age of 24 .\\ 
\hfill 0 &the great olympic gong , born may 6 , 1960 in san antonio , texas , was the first and only indy to win the world champion title of the american\\ 
\hfill 0 &aaron alexander ( born october 27 , 1989 ) is an american professional baseball outfielder for the tampa bay rays of major league baseball {-}lrb\\ 
\hfill 0 &ito's most known work is that of `` ita , the world's best girl '' , an international bestseller written by joão da sampre .\\ 

 \multicolumn{2}{c}{\textbf{\GDC}} \\ 
\hfill 1 &liz carlsson ( born 2 june 1990 ) is a swedish actress and model , most famous for her role as alice in the film ``\\ 
\hfill 0 &{-} `` for other people named john c . white , see john white ( disambiguation ) . '' john c . white , jr . {-}lrb\\ 
\hfill 0 &italo zola ( born 17 june 1959 ) is a former italian footballer . he played as a striker and as a forward for italian clubs pesc\\ 
\hfill 1 &of the year award nominations for 2013 , 2014 and 2015 . her most recent achievement was a `` top 10 debut album '' from her debut album , `` in the name of the devil '' , on\\ 
\hfill 1 &až klimin ( born 20 october 1996 ) is a latvian artistic gymnast . she is a two{-}time european junior team\\ 
\hfill 0 &brian patrick keane ( born may 16 , 1970 ) is an american football defensive end who is currently a free agent . he was drafted by the p\\ 
\hfill 1 &was an english film and television actress . she appeared in many british and american films , and had roles in the tv shows `` my big fat greek wedding '' (\\ 
\hfill 0 &{-} araki ( born january 4 , 1976 in ivanhoe , lautoka ) is a retired brazilian footballer . he played for several clubs\\ 
\hfill 1 &, better known by her stage name pepi , is a korean female singer{-}songwriter . she came to korea after being influenced by kim jin{-}hoon's\\ 
\hfill 1 &( born august 23 , 1962 ) is an american actress . she has appeared in such films as `` kojak '' , `` i saw the fire ''\\ 
\bottomrule 
 \end{tabular} 

    \caption{\textbf{Task 7:} Randomly selected samples from the experiment with \textbf{a single distributional constraint} where $\phi(x) = 1$ iff $x$ contains a mention of a \textbf{female} figure, $\hat{\mu} = 0.5$\label{{table:generation-single}}}
    \end{table*}
    \begin{table*}
    \scriptsize
    \begin{tabular}{p{0.6cm}|p{0.6cm}|p{0.6cm}|p{0.6cm}|p{10cm}}
    \toprule
    $\phi_1(x)$ & $\phi_2(x)$ & $\phi_3(x)$ & $\phi_4(x)$ & \textbf{Sample} \\ 
     \midrule
 \multicolumn{5}{c}{\textbf{\GDCplus}} \\ 
\hfill 0 & 0 & 0 & 1 &, was a russian politician and journalist .\\ 
\hfill 0 & 0 & 0 & 1 &luís alberto herrera carvalho ( born october 6 , 1951 ) is a chilean economist , economist , politician and former mayor of mon\\ 
\hfill 0 & 0 & 0 & 1 &bernard stanton johnson ( born november 8 , 1958 ) is a canadian politician . he was elected to the canadian house of commons in\\ 
\hfill 1 & 0 & 0 & 0 &- > thomas s . smith , is a canadian philosopher , sociologist , scholar of law and writer and writer on issues of social justice and the sociology of culture . smith holds\\ 
\hfill 0 & 0 & 1 & 0 &, known as yuichi takashi , is a japanese professional golfer . takashi was born in shizuoka , japan and attended soto japan golf club\\ 
\hfill 0 & 0 & 0 & 0 &paul r . kelly is a democratic member of the pennsylvania house of representatives . he was elected to represent the 28th legislative district , being reelected in 2006 and 2010 .\\ 
\hfill 1 & 0 & 0 & 1 &sław ( born 12 february 1961 ) is a polish historian , politician , sociologist , and member of the european parliament for poland .\\ 
\hfill 0 & 1 & 0 & 0 &. ( born in dresden , new jersey ) is a german singer and multi{-}instrumentalist who has released several solo albums .\\ 
\hfill 0 & 1 & 0 & 0 &for the artist , see jean{-}luc krüger ( painter ) . '' jean{-}luc krüger ( j\\ 
\hfill 0 & 0 & 1 & 0 &( born april 17 , 1979 in bahrain ) is an iranian footballer who currently plays for al arabi sc .\\ 

 \multicolumn{5}{c}{\textbf{\GDC}} \\ 
\hfill 0 & 0 & 1 & 0 &kim ludwin ( born august 11 , 1985 ) is a canadian ice hockey player who is currently playing with hc slovan bratislava\\ 
\hfill 0 & 1 & 0 & 0 &kazuki shimizu ( born march 30 , 1970 in osaka , japan ) is a japanese mixed martial artist who is the current pride lightweight\\ 
\hfill 0 & 0 & 1 & 0 &andrew jones ( born 23 december 1970 ) is a former english cricketer . jones was a right{-}handed batsman who bowled right{-}\\ 
\hfill 0 & 0 & 1 & 0 &andré fernández de gómez ( born 20 february 1989 ) is a spanish professional footballer who plays for fc barcelona\\ 
\hfill 0 & 0 & 0 & 1 &theodore george hudson ( october 20 , 1877 - april 8 , 1944 ) was a united states army officer . he served as the 19\\ 
\hfill 0 & 0 & 0 & 0 &. he was born in rome , italy on 10 may 1949 .\\ 
\hfill 0 & 0 & 0 & 1 &linda jane thompson ( born march 10 , 1958 ) is an american politician who was the u . s . representative for from 2003 to 2015 .\\ 
\hfill 0 & 1 & 0 & 0 &kenny hansen ( born april 26 , 1982 ) is an american actor best known for his role as the sheriff in the disney channel series `` criminal\\ 
\hfill 0 & 0 & 0 & 1 &in 2007 , he was nominated by the governor of illinois to be the governor of illinois in 2011 for the position of the u . s . representative for illinois's 22nd congressional\\ 
\hfill 0 & 0 & 0 & 0 &the dutch are an influential british reggae music duo , formed in 1982 in dublin . the duo consists of lead vocalist dave schroeder and drummer eric kend\\ 
\bottomrule 
 \end{tabular} 

    \caption{\textbf{Task 8:} Randomly selected samples from the experiment with \textbf{Four distributional constraints}: $\phi_n(x) = 1$ iff $x$ contains at least one of the words from a corresponding $n$-th wordlist proposed by \citep{plug_and_play_20}. The considered wordlists are ``science", ``art", ``sports" and ``business" and for each $\hat{\mu}_n = 0.25$\label{{table:generation-mult-dist}}}
    \end{table*}
    \begin{table*}
    \scriptsize
    \begin{tabular}{p{0.6cm}|p{0.6cm}|p{12cm}}
    \toprule
    $\phi_1(x)$ & $\phi_2(x)$ & \textbf{Sample} \\ 
     \midrule
 \multicolumn{3}{c}{\textbf{\GDCplus}} \\ 
\hfill 1 & 1 &; ( born 10 october 1987 ) is an iranian footballer who plays as a defender for bursaspor and the iran national football team . she is\\ 
\hfill 1 & 1 &. she is the daughter of vladimir uchadze , who is also a former russian football player .\\ 
\hfill 0 & 1 &kenzo shiro ( born 26 april 1985 ) is a japanese football player who currently plays for j . league division 2 club japanese super\\ 
\hfill 0 & 1 &hans schuetke ( born 21 july 1953 ) is a german former footballer who played as a forward for vfb stuttgart , sheffield\\ 
\hfill 0 & 1 &, real name marc valera cipriles ( born 4 may 1969 ) is a former costa rican footballer who last played as a defender .\\ 
\hfill 0 & 1 &brent lincoln ( born 1 october 1985 ) is an english footballer who plays as a striker for bristol rovers . born in bristol , lincoln\\ 
\hfill 0 & 1 &joseph e . `` joey '' bierer ( born may 18 , 1953 in columbus , ohio ) is a retired american basketball player\\ 
\hfill 0 & 1 &aryeh ( ; born 22 october 1988 ) is an israeli footballer currently playing for kfar saba .\\ 
\hfill 0 & 1 &juan de almagro castro ( born 21 october 1981 in lisbon ) is a portuguese retired footballer who played as a midfielder . he\\ 
\hfill 1 & 1 &is a canadian tennis player . as of 2014 , she has a wta singles career high ranking of 967 achieved on july 15 , 2015 .\\ 

 \multicolumn{3}{c}{\textbf{\GDC}} \\ 
\hfill 0 & 1 &sébastien lépine ( born 9 march 1987 ) is a french football player currently playing for olympique lyonnais in ligue 1 .\\ 
\hfill 1 & 0 &in a career that spans nearly four decades , león has starred in some of the most successful movies of the late{-}1980s and early{-}1990s . her breakthrough came in the 2005 film\\ 
\hfill 0 & 1 &hamed sargam ( born 9 january 1975 ) is a saudi arabian footballer . he played for al qadisiyah in saudi ar\\ 
\hfill 0 & 1 &james `` jim '' mcgrath ( born may 24 , 1934 ) is a former professional american football player who played wide receiver for eight seasons for the\\ 
\hfill 0 & 1 &james `` jack '' lancaster ( born 21 march 1935 ) is an english former footballer who played in the football league for brentford , leeds united\\ 
\hfill 0 & 1 &aacson de rosas de lópez , jr . ( born 18 april 1976 in barcelona ) is a spanish professional racing cyclist .\\ 
\hfill 1 & 1 &, born on 29 april 1982 in baku ) is a professional turkish tennis player . she reached her highest wta singles ranking of 280 on 20 september 2012 .\\ 
\hfill 1 & 1 &'( , born september 10 , 1992 ) is a female water polo player of kenya . she was part of the kenyan team at\\ 
\hfill 0 & 1 &( november 10 , 1981 in davao ) is a dutch footballer who plays for vitesse as a defender .\\ 
\hfill 1 & 1 &, born november 15 , 1986 in tokyo , japan ) is a japanese volleyball player . she was drafted fifth in the 2011 j . league division 1\\ 
\bottomrule 
 \end{tabular} 

    \caption{\textbf{Task 9:} Randomly selected samples from the experiment with a \textbf{hybrid distributional constraint} where $\phi_1(x) = 1$ iff $x$ contains a mention of a \textbf{female} figure, $\hat{\mu}_1 = 0.5$ and $\phi_2(x) = 1$ iff $x$ contains at least one of the words from the \textbf{``sports" wordlist} proposed by \citep{plug_and_play_20} and $\hat{\mu}_2 = 1$\label{{table:generation-hybrid-science}}}
    \end{table*}
    \begin{table*}
    \scriptsize
    \begin{tabular}{p{0.6cm}|p{0.6cm}|p{12.8cm}}
    \toprule
    $\phi_1(x)$ & $\phi_2(x)$ & \textbf{Sample} \\ 
     \midrule
 \multicolumn{3}{c}{\textbf{\GDCplus}} \\ 
\hfill 1 & 1 &, born 3 may 1947 in selju , turkey ) is a former turkish women's football player . she was a student in istanbul , istanbul .\\ 
\hfill 1 & 1 &is a french filmmaker and academic . she is known for her documentary , `` le seigneur une réunion de bahaudouin '' , the first in which a french student walks around\\ 
\hfill 1 & 1 &, also known by her married name , was a japanese scientist and a scientist who specialized in nuclear physics and nuclear radiation . she was the second woman , after kumiko ouchi ,\\ 
\hfill 1 & 1 &was an indian historian and scholar in the field of indian history . she is known for her book `` sanskrit , chakri and kanchra '' ( 18\\ 
\hfill 0 & 1 &, born on april 24 , 1957 , in chungzhou , shandong , was a chinese politician and academic who served as a member of the legislative yuan from july 12 ,\\ 
\hfill 0 & 1 &, ( ; january 26 , 1917 - may 6 , 1997 ) was a russian politician , scientist , and diplomat . from the early 1930s to the mid\\ 
\hfill 0 & 0 &israel hanadiyev ( ; born april 8 , 1985 ) is a russian{-}born russian professional football player . he plays for fc\\ 
\hfill 1 & 1 &linda borregoni is an american astronomer and theoretical cosmologist . she has received numerous awards , including a macarthur foundation fellow for astronomy award for her work in cosmology\\ 
\hfill 0 & 1 &sarah c . lee ( born january 25 , 1931 ) is an american educator , academic and medical researcher . lee has written a series of books\\ 
\hfill 0 & 1 &alexander leonard bernstein ( born 8 april 1940 in breslau , switzerland ) is a swiss nuclear scientist and politician who\\ 

 \multicolumn{3}{c}{\textbf{\GDC}} \\ 
\hfill 0 & 1 &: ( 1558 - 7 june 1628 ) was a french writer , philosopher , journalist , antiquary , lawyer and historian . he was one of the great\\ 
\hfill 1 & 0 &, was an ancient egyptian princess . she was the daughter of the egyptian empress nikhaït of zagros .\\ 
\hfill 1 & 1 &saysia nand is a student of asean university and sri lanka university of science and technology and her doctoral student is shahid srinivasan . nand has\\ 
\hfill 0 & 1 &b{-} ( born may 26 , 1977 ) is a canadian historian , and former chair of the department of medieval history of the university of british columb\\ 
\hfill 1 & 1 &sara sara ( born july 3 , 1954 ) is an american social scientist . she is a co{-}director of the national center for family research and\\ 
\hfill 0 & 1 &: born 13 october 1969 ) is a british philosopher . he is professor of philosophy at the university of london and chair of the department of philosophy of humanistic philosophy\\ 
\hfill 0 & 1 &, was a chinese poet , playwright , translator , translator , sociologist and academic . he was born in sichuan in 1796 and became an early member of the literary association of\\ 
\hfill 0 & 1 &larry t . ellerbe is an american scientist who is the founding director of the department of natural resources and environment at the carnegie mellon university . he is the son of the\\ 
\hfill 0 & 0 &a . p . taylor is an american professor of philosophy and director of the department of philosophy of religion at the university of california , berkeley . his recent research has focused on\\ 
\hfill 0 & 1 &, was an israeli arabologist , historian , and scholar of early israel . he is best known as the former director of the national library of the israel .\\ 
\bottomrule 
 \end{tabular} 

    \caption{\textbf{Task 10:} Randomly selected samples from the experiment with a \textbf{hybrid distributional constraint} where $\phi_1(x) = 1$ iff $x$ contains a mention of a female figure, $\hat{\mu}_1 = 0.5$ and $\phi_2(x) = 1$ iff $x$ contains at least one of the words from the \textbf{``science" wordlist} proposed by \citep{plug_and_play_20} and $\hat{\mu}_2 = 1$\label{{table:generation-hybrid-sports}}}
    \end{table*}
    
\end{document}